# Atom Search Optimization with Simulated Annealing – a Hybrid Metaheuristic Approach for Feature Selection

*Kushal Kanti Ghosh, Ritam Guha, Soulib Ghosh, Suman Kumar Bera\*, Ram Sarkar*
*Department of Computer Science and Engineering, Jadavpur University, Kolkata, India*
*[kushalkanti1999@gmail.com](kushalkanti1999@gmail.com), [ritamguha16@gmail.com](ritamguha16@gmail.com), [ghoshsoulib@gmail.com](ghoshsoulib@gmail.com),*
*[berasuman007@gmail.com](berasuman007@gmail.com), [ramjucse@gmail.com](ramjucse@gmail.com)*

**Abstract:** *'Hybrid meta-heuristics' is one of the most interesting recent trends in the field of optimization and feature selection (FS). In this paper, we have proposed a binary variant of Atom Search Optimization (ASO) and its hybrid with Simulated Annealing called ASO-SA techniques for FS. In order to map the real values used by ASO to the binary domain of FS, we have used two different transfer functions: S-shaped and V-shaped. We have hybridized this technique with a local search technique called, SA We have applied the proposed feature selection methods on 25 datasets from 4 different categories: UCI, Handwritten digit recognition, Text/non-text separation, and Facial emotion recognition. We have used 3 different classifiers (K-Nearest Neighbor, Multi-Layer Perceptron and Random Forest) for evaluating the strength of the selected featured by the binary ASO, ASO-SA and compared the results with ≥4 recent wrapper-based algorithms. The experimental results confirm the superiority of the proposed method both in terms of classification accuracy and number of selected features.*

**1. Introduction**

In recent times, due to the wide availability of data and the crucial need to extract useful information from these data, in information industry data mining is one of the fastest growing research topics (M. Mafarja et al., 2018). Data mining is an integral part of knowledge discovery (M. M. Mafarja & Mirjalili, 2017), which consists of data pre-processing, data mining, pattern evaluation and knowledge representation. The pre-processing step has a huge impact on the data mining technique. Now, all the extracted features are not useful for data mining, as some of those are irrelevant or redundant (Ghaemi & Feizi-Derakhshi, 2016), which affects the data mining tasks (e.g., classification). Not only that these irrelevant or redundant features become burden in process of knowledge discovery, but also they result in increased computational complexity, incomprehensibility of the results and augmented storage requirement (Ghaemi & Feizi-Derakhshi, 2016). Here comes the importance of feature selection (FS) techniques. FS (Anaraki & Usefi, 2019) is a pre-processing step that helps in dimensionality reduction of the dataset being processed by eliminating the irrelevant/redundant features. FS is a proper subset of feature weighting (Wettschereck, Aha, & Mohri, 1997), in which a feature is assigned a weight according to their importance in solving a problem using some machine learning approach. FS is subset of feature weighting, where weights are either '1' or '0', i.e., a feature is either selected or rejected.

Coarsely, the main steps of FS (DASH & LIU, 1997) are: subset generation, subset evaluation, setting end criteria and validation. Based upon the subset evaluation criteria, FS methods can be divided in two types: filter and wrapper. Wrapper methods use a learning algorithm to measure the worth of a feature subset (M. Ghosh, Adhikary, et al., 2019). In filter methods, no learning algorithm is used, instead, the importance of a feature or a subset of features is determined based on some specific characteristics of the dataset under consideration (M. Ghosh, Adhikary, et al., 2019).

The subset generation is basically a searching technique that selects a subset from the original set using complete, random or heuristic search. In complete search, all possible subsets are evaluated to find the best one. Though this process guarantees the finding of best feature subset, for n number of features we need to evaluate $2^n$ subsets. Obviously the computational complexity of this method is too high, thus making a complete search non-feasible option (M. Ghosh, Adhikary, et al., 2019). Random search is another strategy for generating feature subset (M. Mafarja et al., 2018), but in worst case it may perform an exhaustive search. Heuristic search is another alternative to the mentioned search strategies. It can also be called depth first search guided by heuristics (M. Mafarja et al., 2018). Various meta-heuristics such as Genetic Algorithm (Yang & Honavar, 1998), Ant Colony Optimization (ACO) (Dorigo & Di Caro, n.d.), Particle Swarm Optimization (PSO) (Lee, Soak, Oh, Pedrycz, & Jeon, 2008), Dragonfly Algorithm (M. Mafarja, Heidari, Faris, Mirjalili, & Aljarah, 2020), Grasshopper Optimization Algorithm (GOA) (Saremi, Mirjalili, & Lewis, 2017), Forest Optimization Algorithm (Ghaemi & Feizi-Derakhshi, 2016), Grey Wolf Optimizer (GWO) (Emary, Zawbaa, Grosan, & Hassenian, 2015), etc. have shown quite good results in tackling FS problems. Meta-heuristic algorithms can be classified mainly in two categories (M. Mafarja et al., 2018): single solution and population based algorithms. In single solution-based methods, one state is manipulated and transformed throughout the search process. Methods in this category show more exploitative nature, which implies focusing on the space around a possible solution. In population-based algorithms, a set of solutions is evolved during the search process. These methods are more explorative in nature, which means looking around different regions of the space. While designing a meta-heuristic algorithm, both of these criteria must be considered. High exploitation may result in the optimizer being trapped in local optima whereas high exploration may lead to selection of outliers.

Atom Search Optimization (ASO) (Zhao, Wang, & Zhang, 2019b) is the newest addition to the domain of meta-heuristic algorithms. ASO tries to mimic the movement of atoms according to Lennard-Jones potential (Stone, 2013) developed among interacting atoms. In recent past, the algorithm has gained popularity as a nature-inspired optimization algorithm and has been applied to solve various interesting optimization problems like hydro-geologic parameter estimation (Zhao et al., 2019b), dispersion coefficient estimation (Zhao, Wang, & Zhang, 2019a) etc. But to the best of our knowledge, the optimization procedure used in ASO has not ever been applied to solve any FS problems. But the very characteristics of ASO makes it a suitable choice for FS problems. This has motivated us to develop the version of ASO which is applicable to FS. In addition, to improve the exploitation ability of the basic ASO, we have combined a popular local search technique called Simulated Annealing (SA) with the existing ASO version. Finally, to prove the robustness and applicability of ASO in the FS domain, we have made a case study by applying our ASO-SA based FS model over varied datasets which include UCI, text/non-text classification, digit recognition and facial emotion recognition.

In this paper, we have presented our experimental findings regarding the newly developed FS version of ASO. Our primary contributions to this paper are listed below:

- Development of the FS version of ASO for the first time.
- Hybridization of ASO with SA denoted by ASO-SA to improve exploitation ability of basic ASO.
- A detailed case study to show the usefulness of ASO and ASO-SA in the FS domain by applying them over UCI, text/non-text classification, digit recognition and facial emotion recognition datasets.

The rest of the paper is organized as follows: Section 2 provides a brief overview of the other works done by researchers in the domain of FS. Section 3 describes the proposed work in detail. The results and findings of our case study are presented in section 4. Section 5 concludes the paper and provides the scope of future works.

## 2. Related Work

Over the years, researchers have found that simple natural phenomena have gifted abilities to provide solutions to many hard optimization problems. Motivated from these findings, people have implemented and used various nature and bio-inspired meta-heuristic algorithms to solve the optimization problems. One such highly addressed optimization problem is FS where researchers have proposed many algorithms to search for a near-optimal subset of features within a limited time span.

One of the most fundamental optimization algorithms applied to solve FS problems is Genetic Algorithm (GA). After its inception as an optimization algorithm (Holland, 1992), GA has been used to solve multiple optimization problems like gene selection in cancer classification (Alba, Garcia-Nieto, Jourdan, & Talbi, 2007), stock market data mining optimization (Lin, Cao, Wang, & Zhang, 2004), component designs (Husbands, Jermy, McIlhagga, & Ives, 1996) etc. A more comprehensive set of applications of GA can be found in (Tang, Man, Kwong, & He, 1996). After the introduction of GA to the domain of FS by Leardi et al. in (Leardi, 1996), it has been used extensively to solve various FS problems (Basiri & Nemati, 2009; De Stefano, Fontanella, Marrocco, & Di Freca, 2014; Il-Seok Oh, Jin-Seon Lee, & Byung-Ro Moon, 2004; Malakar, Ghosh, Bhowmik, Sarkar, & Nasipuri, 2019; Nemati, Ehsan, Ghasem-aghaee, & Hosseinzadeh, 2009; Prasad, Biswas, & Jain, 2010). GA uses the concepts of chromosome crossover and mutation to achieve a balance in exploration and exploitation of the search space. Inspired from the success of GA in the FS domain, researchers have made efforts to adapt other popular metaheuristic optimization algorithms to FS. As a result, some highly used optimization algorithms like PSO (Chen, Zhou, & Yuan, 2019; C. Huang & Dun, 2008; Wang, Yang, Teng, Xia, & Jensen, 2007; Xue, Zhang, & Browne, 2012), ACO (Basiri & Nemati, 2009; Forsati, Moayedikia, Jensen, Shamsfard, & Meybodi, 2014; C. L. Huang, 2009; Kabir, Shahjahan, & Murase, 2012; Kashef & Nezamabadi-pour, 2015), GSA (Papa et al., 2011; E. Rashedi & Nezamabadi-Pour, 2014; Esmat Rashedi & Nezamabadi-pour, 2012; Esmat Rashedi, Nezamabadi-Pour, & Saryazdi, 2010b) has been re-introduced to the research community with their binary versions to solve FS problems. These binary versions have performed well in the domain of FS and have become quite famous in the research community. But each such algorithm suffers from a fundamental problem i.e. finding a proper trade-off for exploration and exploitation. The next generation of researchers have made efforts to solve this problem by hybridizing various algorithms together (Anter, Azar, & Fouad, 2019; Badawi, Khalil, & Alsmadi, 2013; Trivedi, Srivastava, Mishra, Shukla, & Tiwari, 2018). Although some algorithms have performed better than the others to find a proper balance between exploitation and exploration, the problem still remained. In the search of a solution to this problem, researchers have started to exploit the searching behaviour of other natural objects and tried to mimic them which resulted in the introduction of some recently proposed metaheuristic algorithms in FS like Whale Optimization Algorithm (WOA) (Hussien, Hassanien, Houssein, Bhattacharyya, & Amin, 2019; M. M. Mafarja & Mirjalili, 2017; Sharawi, Zawbaa, & Emary, 2017), GWO (Abdel-Basset, El-Shahat, El-henawy, de Albuquerque, & Mirjalili, 2020; Emary, Zawbaa, & Hassanien, 2016), GOA (M. Mafarja et al., 2019), Artificial Bee Colony (ABC) (Zhang, Cheng, Shi, Gong, & Zhao, 2019), Dragonfly algorithm (M. M. Mafarja, Eleyan, Jaber, Hammouri, & Mirjalili, 2017) etc. The applicability of these algorithms in the FS domain have inspired us to develop a binary version of a recently proposed optimization technique named ASO and to apply it over various datasets.

Zhao et al. proposed the ASO model in (Zhao et al., 2019a) to estimate dispersion coefficient in ground water. Through the proposed model, the authors have tried to mimic the characteristics of atomic motion. According to the concepts of dynamics of atoms, Lennard-Jones potential and bond-length potential among atoms produce the interaction forces among atoms which guides their movements. Using these concepts, the ASO model has been developed to provide solution to optimization problems. The authors have applied ASO for dispersion coefficient estimation of ground water. The experimental outcomes prove that ASO is able to outperform other popular optimization approaches like PSO, GSA and Bacterial Foraging Optimization (BFO) (Passino, 2010). Zhao et al. have used the same procedure again to solve hydrogeological parameter estimation problem in (Zhao et al., 2019b). In this paper, they have provided the experimental outcomes of the application of ASO over 37 benchmark unimodal, multimodal, hybrid, low-dimensional and composite functions. The results obtained by ASO have been tallied against PSO, GA, SA, GSA and Wind Driven Optimization (WDO) (Bayraktar, Komurcu, Bossard, & Werner, 2013). The excellent results of ASO clearly establishes its applicability in the optimization domain. But ASO has never been used to solve FS problems. The searching prowess of ASO has motivated us to apply it to search for important feature subsets from feature sets in various contexts. That is why we have developed a binary version of ASO with the help of transformation functions to map real values in ASO to binary values. In addition, to prove the robustness of this version of ASO, we have applied it over datasets and features with different backgrounds.

## 3. Present Work

### 3.1 Brief Overview of Atom Search Optimization (ASO)

ASO is inspired by basic molecular dynamics. Molecular dynamics was first proposed in the field of theoretical physics (Alder & Wainwright, 1959) but afterwards it has been used also in other fields of science like biology, chemistry and materials science. The atomic motion follows the classical mechanics (Goldstein, Poole, Safko, & Addison, 2002). According to second law of Newton, if $F_i$ is the interaction force and $G_i$ is the constraint force working on the $i^{th}$ atom and the atom has mass $m_i$ then the acceleration of the atom is (Ryckaert, Ciccotti, & Berendsen, 1977):

$$a_i = \frac{F_i + G_i}{m_i} \quad (1)$$

### 3.1.1 Interaction Force

Lennard-Jones (LJ) potential (Stone, 2013) is a simple mathematical model that approximates the interaction force between a pair of atoms. The LJ potential between $i^{th}$ and $j^{th}$ atom is given by

$$U(r_{ij}) = 4\varepsilon\left[\left(\frac{\sigma}{r_{ij}}\right)^{12} - \left(\frac{\sigma}{r_{ij}}\right)^{6}\right] \quad (2)$$

Where, $\varepsilon$ is the depth of the potential well (Zhao et al., 2019b), $\sigma$ is the finite distance at which the inter-particle potential is zero, $r_{ij} = x_j - x_i$ where $x_i = (x_{i1}, x_{i2}, \ldots, x_{in})$ and $x_j = (x_{j1}, x_{j2}, \ldots, x_{jn})$ are the positions of ith and jth atom in the nth dimension. So the Euclidian distance between $x_i$ and $x_j$ is given by $r_{ij} = \|x_j - x_i\| = \sqrt{(x_{i1} - x_{j1})^2 + (x_{i2} - x_{j2})^2 + \cdots + (x_{in} - x_{jn})^2}$. The terms $\left(\frac{\sigma}{r_{ij}}\right)^{12}$ and $\left(\frac{\sigma}{r_{ij}}\right)^{6}$ represent attraction and repulsion respectively. Figure 1 explains the nature of this LJ potential.

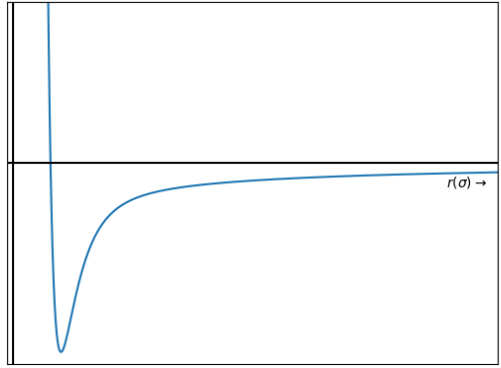

Figure 1: LJ potential curve with attraction and repulsion region. In the attractive zone, the attraction force gradually decreases to zero as the distance between the atoms increases. In the repulsive zone, the repulsive force rapidly increases as the distance decreases. At the equilibrium distance (r=1.2σ), the interaction force between the atoms is zero. At that point, the minimum bonding potential energy of the atoms is reached.

The interaction force working on $i^{th}$ atom from $j^{th}$ atom in the $d^{th}$ dimension at the $t^{th}$ time is given by (Zhao et al., 2019a):

$$F_{ij}^d(t) = -\nabla U(r_{ij}) = \frac{24\varepsilon(t)}{\sigma(t)}\left[2\left(\frac{\sigma(t)}{r_{ij}(t)}\right)^{13} - \left(\frac{\sigma(t)}{r_{ij}(t)}\right)^{7}\right]\frac{r_{ij}}{r_{ij}^d} \quad (3)$$

And

$$F'_{ij}(t) = \frac{24\varepsilon(t)}{\sigma(t)}\left[2\left(\frac{\sigma(t)}{r_{ij}(t)}\right)^{13} - \left(\frac{\sigma(t)}{r_{ij}(t)}\right)^{7}\right] \quad (4)$$

The atom keeps a relative distance (from each other) that varies in a certain range all the time due to attraction or repulsion, and the change of amplitude in repulsion relative to equilibrium distance is much greater than that in attraction. Repulsion is positive and attraction is negative, so the atoms cannot converge to a specific position. Equation (4) cannot be directly used for optimization purpose. So the authors of (Zhao et al., 2019b) have proposed a modified version of this for optimization problems:

$$F'_{ij}(t) = -\eta(t)\left[2\bigl(h_{ij}(t)\bigr)^{13} - \bigl(h_{ij}(t)\bigr)^{7}\right] \qquad (5)$$

where η(t) is depth function to adjust the attraction or repulsion region. It is defined as:

$$\eta(t) = \alpha\left(1 - \frac{t-1}{T}\right)^{3} e^{-\frac{20t}{T}} \qquad (6)$$

where T is the maximum number of iteration and α is the depth weight.

$$h_{ij}(t) = \begin{cases} h_{min} & \text{if } \dfrac{r_{ij}(t)}{\sigma(t)} < h_{min} \\ \dfrac{r_{ij}(t)}{\sigma(t)} & \text{if } h_{min} \le \dfrac{r_{ij}(t)}{\sigma(t)} \le h_{max} \\ h_{max} & \text{if } \dfrac{r_{ij}(t)}{\sigma(t)} > h_{max} \end{cases} \qquad (7)$$

$h_{min}$ and $h_{max}$ are lower limit and upper limit of $h_{ij}$ respectively. The length scale σ(t) is defined as $\sigma(t) = \left\| x_{ij}(t), \frac{\sum_{j \in KBest} x_{ij}(t)}{K(t)} \right\|$, where KBest is a subset of K atoms with the best function fitness values.

$$\begin{cases} h_{min} = g_0 + g(t) \\ h_{max} = u \end{cases} \qquad (8)$$

where u is the upper limit and g is a drift function that helps the algorithm to drift from exploration to exploitation and is given by:

$$g(t) = 0.1 \times \sin\left(\frac{\pi}{2} \times \frac{t}{T}\right) \qquad (9)$$

Now the total force working on the ith atom from all other atoms is the weighted sum of components of the forces in dth dimension:

$$F_i^d(t) = \sum_{j \in KBest} random_j F_{ij}^d(t) \qquad (10)$$

where $random_j$ is a random number in [0,1].

### 3.1.2 Geometric Constraints:

In the motion of atoms, the geometric constraint in molecular dynamics plays a vital role. For simplicity, in ASO an atom is assumed (Zhao et al., 2019b) to have a covalent bond with the best atom, so each atom is acted upon by a constraint force by the best atom. This constraint force acting on the $i^{th}$ atom in the $d^{th}$ dimension is defined as:

$$G_i^d = \lambda(t)\left(x_{best}^d(t) - x_i^d(t)\right) \tag{11}$$

where, $\lambda(t)$ is the Lagrangian multiplier, defined as

$$\lambda(t) = \beta e^{-\frac{20t}{T}} \tag{12}$$

where, $\beta$ is the multiplier weight.

### 3.1.3 Atomic Motion:

Considering both interaction force and geometric constraints, the acceleration of the $i^{th}$ atom at time t in $d^{th}$ dimension is

$$\begin{aligned}a_i^d(t) =\;& \frac{F_i^d(t) + G_i^d(t)}{m_i} \alpha \left(1 - -\frac{t-1}{T}\right)^3 e^{\frac{-20t}{T}} \\ & * \sum_{j \in KBest} \frac{random_j \left[2(h_{ij}(t))^{13} - (h_{ij}(t))^7\right](x_j^d(t) - x_i^d(t))}{m_i(t)} \frac{}{\|x_i(t), x_j(t)\|_2} \\ & + \beta e^{-\frac{20t}{T}} \frac{x_{best}^d(t) - x_i^d(t)}{m_i(t)}\end{aligned} \tag{13}$$

$m_i(t)$ is the mass of the $i^{th}$ atom at $t^{th}$ iteration. It is determined as:

$$M_i(t) = e^{-\frac{Fit_i(t) - Fit_{best}(t)}{Fit_{worst}(t) - Fit_{best}(t)}} \tag{14}$$

$$m_i(t) = \frac{M_i(t)}{\sum_{j=1}^{N} M_j(t)} \tag{15}$$

Where $Fit_{best}(t) = \min_{i=\{1,2,\ldots,n\}} Fit_i(t)$ and $Fit_{worst}(t) = \max_{i=\{1,2,\ldots,n\}} Fit_i(t)$

The position and velocity of the $i^{th}$ atom at time $(t + 1)$ are given by:

$$v_i^d(t+1) = random_i^d v_i^d(t) + a_i^d(t) \tag{16}$$

$$x_i^d(t+1) = x_i^d(t) + v_i^d(t+1) \tag{17}$$

In order to apply it to the optimization problem, exploration ability needs to be enhanced. Hence in the beginning, each atom needs to interact with large number of atoms with better fitness values, as its neighbors, so the value of K must be high. To enhance exploitation towards the ending of the algorithm, number of neighbors K must be decreased. So, K is calculated as:

$$K(t) = N - (N - 2)\sqrt{\frac{t}{T}} \qquad (18)$$

**3.2 Binary ASO for feature selection:**

In this section, we describe the customizations we made to use ASO for FS. The proposed methodology is wrapper-based FS method. In wrapper methods, the subsets of features are evaluated using a classifier to select the optimum subset. We have used ASO as the search method to find the optimal subset from the original feature set. We have used Multi-layer Perceptron, Random-Forest and K-Nearest Neighbor classifiers for the subset evaluation purpose. To use ASO for FS purpose, we need to define two things accordingly: representing an atom and defining a fitness function.

**3.2.1 Atom representation:**

FS is considered as a binary optimization problem, i.e., solutions are restricted to binary values. To represent a FS problem, we need a vector of 0s and 1s, where 0 represents the corresponding feature is not selected and 1 represents selected. The vector length is equal to the number of features in the original dataset. In this case, a binary feature vector is considered as an atom.

**3.2.2 Fitness function:**

The fitness function consists of classification error and number of selected features. This justifies the fact that FS can be considered a multi-objective problem that aims to find smallest subset with best accuracy. The fitness function is defined as:

$$\downarrow \text{Fitness} = \omega \times \varepsilon + (1 - \omega) \times \frac{|X|}{|D|} \qquad (19)$$

$|X|$ and $|D|$ represent respectively number of selected features and total number of features, $\varepsilon$ is the classification error rate of the corresponding classifier and $\omega \epsilon [0,1]$ is the weight.

Since in FS problem the knowledge representation is binary in nature, the atoms can move by flipping various numbers of bits. So, we need to modify the position updating process. In continuous version of ASO, the atoms can move around the search space utilizing the position vectors within the continuous real domain. So, position can be easily updated by adding velocity using equation (17). But in binary space, since we deal with only 0 and 1, the same formula to update position will not work. So, we need a way to map velocity and position to fit into binary positions. Now, according to (Kennedy & Eberhart, 1997), we can change the position of an atom based on the probability of velocity of that atom. To do this, a transfer function is required to map velocity values to probability values. After converting velocity values to probability values, we can update the position vectors using the corresponding probability of velocities as:

$$x_i^d(t+1) = \begin{cases} \neg x_i^d(t) & \text{if rand} < T(v_i^d(t+1)) \\ x_i^d(t) & \text{if rand} \geq T(v_i^d(t+1)) \end{cases} \quad (20)$$

where, T is the transfer function. For this work we have set rand = 0.5. Actually, transfer functions force atoms to move in a binary space (Mirjalili & Lewis, 2013). According to (Esmat Rashedi, Nezamabadi-Pour, & Saryazdi, 2010a), to select a transfer function for mapping velocity values to probability values some points must be maintained:

1. Since transfer functions indicate probability, the range must be [0,1]
2. For a large absolute value of the velocity, transfer function should output high probability for position change (0 to 1 or vice-versa). Similarly, for small absolute value of the velocity, the output should be low.
3. With the increase of velocity, the value of the transfer function must rise. If an atom is moving away from the best solution, it must have a higher probability of changing its position to previous one. Similarly, with the decrease of the velocity the value of the transfer function must decrease.

In this work, we have used two different types of transfer functions (Mirjalili & Lewis, 2013):

1. S-shaped Transfer Function:

$$T(x) = \frac{1}{1 + e^{-x}}$$

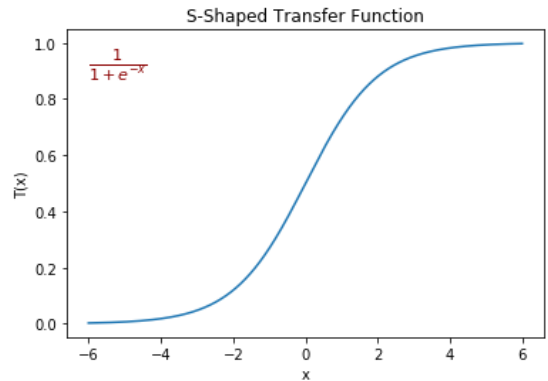

Figure 2: S-shaped transfer function

2. V-shaped Transfer Function:

$$T(x) = |\tanh(x)|$$

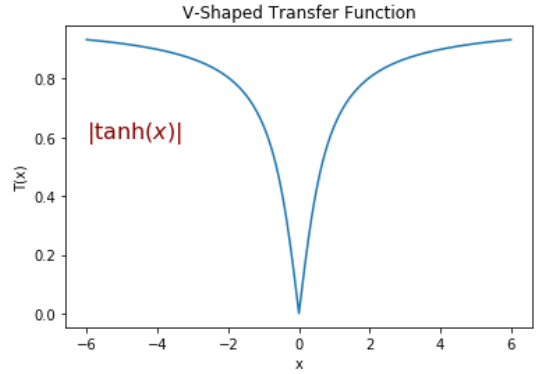

Figure 3: V-shaped transfer function

### 3.2.2 Parameters of ASO:

As parameters, ASO algorithm needs: number of atoms in a population, maximum number of iterations and dimension of the problem. In FS, this dimension is the number of features representing the corresponding dataset. In equation (2), we set $u = 1.24$ following the work (Zhao et al., 2019a). The other parameters to be determined are depth (α) and multiplier weight (β). Following the works of (Zhao et al., 2019b), we set α and β as α = 50 and β = 0.2 for our experiments. We have set number of atoms in the population as 20 and maximum number of iterations as 30.

### 3.3 Simulated Annealing-based ASO (ASO-SA):

One of the main challenges in FS is to search for a proper trade-off between exploration and exploitation. Each FS algorithm has its own way to find such a balance. In ASO, this trade-off is managed by the value of K(t) which is computed by equation (16). In order to ensure exploration at the initial stage, the value of K(t) is kept large (for small t values). Large value of K(t) lets each atom interact with many of its neighbors having better fitness values. With increasing iteration (t), the value of K(t) decreases and the atoms interact with a smaller number of its neighbors having better fitness values which, in turn, enhances exploitation in the later stages. Thus, ASO ensures exploration during initial iterations and exploitation during final iterations. But, the value of K(t) may become as low as 2 which cannot lead to proper exploitation of the search space and makes it almost impossible to avoid local optima. To increase the exploitation ability of ASO and bypass the stagnation of atoms in local optima, we have embedded a local search technique named Simulated Annealing (SA) into ASO. Proposed by Kirkpatric et al. in (Kirkpatrick, Gelatt, & Vecchi, 1983), SA is hill climbing based single-solution meta-heuristic procedure which accepts worse solutions with some probability in the hope to circumvent local optima. After each iteration of ASO, each atom in the final set of population undergoes SA to improve the quality of the solution and to overcome local optima. The algorithm begins with creating a neighbor of an atom through perturbation of the atom. If the fitness of the neighbor exceeds that of the atom, then the neighbor is always accepted as a solution else it is accepted with certain probability determined by Boltzmann probability which is defined in equation (21) where curFitness is the fitness of the generated neighbor and bestFitness is the fitness of the best solution found so far among all the generated neighbors. Temp is an important parameter in the process known as the temperature which determines the number of times neighbors needs to be evaluated for a particular atom. Temp is initialized to $2 * |N|$ where $|N|$ is the total number of features present in the dataset and decreased as $Temp = 0.93 * Temp$ (cooling schedule) as used in (Jensen & Shen, 2003).

$$\text{Boltzmann probability } p = e^{\frac{-(curFitness - bestFitness)}{Temp}} \quad (21)$$

In this way, we have made an attempt to combine the concepts of SA with ASO to improve the exploitation ability of ASO.

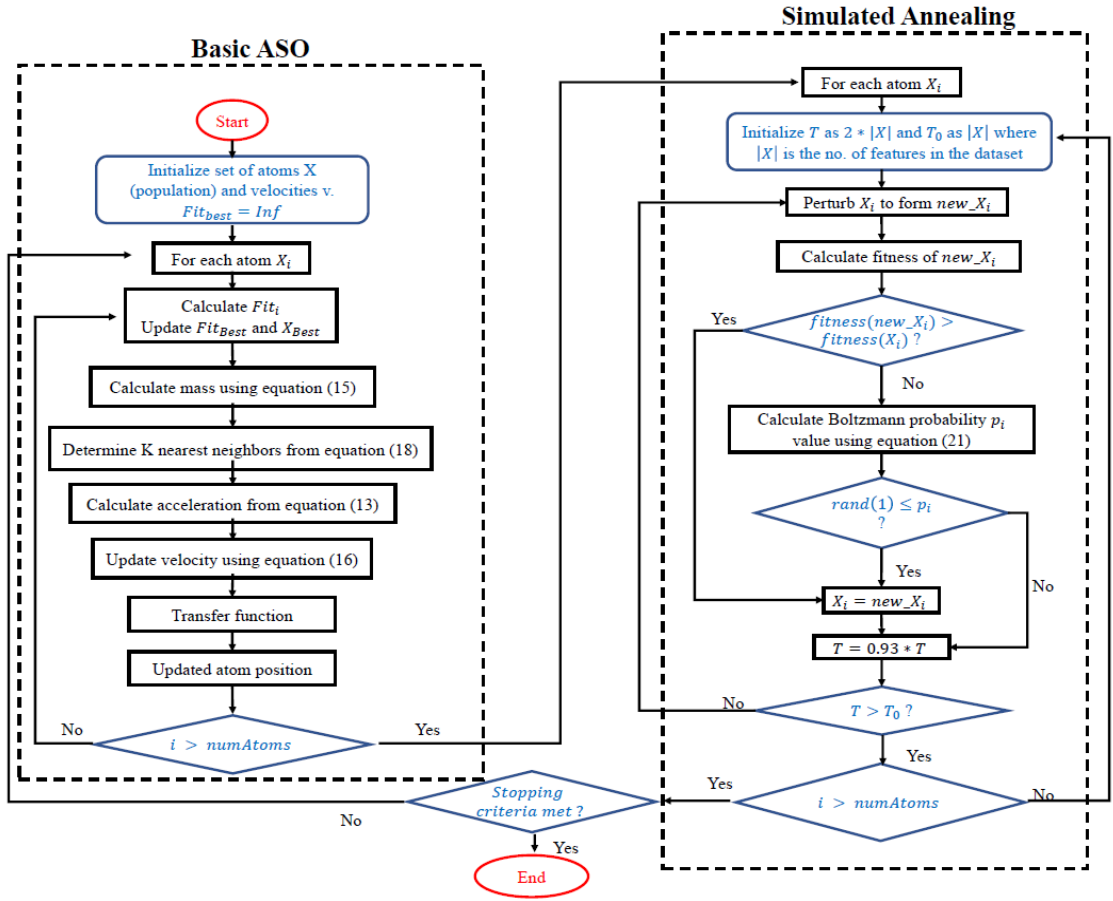

Figure 4: The entire flowchart of ASO-SA. The Basic ASO and Simulated Annealing portions are marked separately with dotted lines

The entire flow of the ASO-SA algorithm is presented in Figure 4. The basic ASO and SA are separately marked. These two processes interact with each other to form solutions of ASO-SA.

## 4. Experimentation

In order to evaluate the effectiveness of the proposed FS method, we have applied it to different pattern classification problems. We have shown how our proposed method ensures the same accuracy (as obtained with the original feature vector), if not better, using much smaller number of features selected by the FS algorithm. Again, to prove its effectiveness, we have compared it with other well-known FS algorithms. We have used four case studies considering four different categories of pattern classification problems – UCI data, handwritten digit recognition, text non-text separation and FER. All the case studies, dataset descriptions and the feature extraction methods are explained in the following subsections.

**4.1 Case Study:**

Here, we have described the datasets we have used and the corresponding feature extraction techniques. It is to be noted that for UCI datasets, feature sets are already given, but for the rest, we have applied some standard feature extraction methods.

**4.1.1 Case study 1: UCI datasets**

Seven different benchmark datasets from UCI repository (Dua & Graff, 2017), have been used to evaluate the proposed FS method. The datasets are described in terms of number of features, number of instances, number of classes and application domain in Table 1. For these datasets, we have used K-Nearest Neighbor (K = 5) classifier as an evaluator in the wrapper FS framework. For each dataset, 80% of the instances are used for training and rest 20% used for testing following the works found in the literature (M. Mafarja et al., 2019)(Alsaafin & Elnagar, 2017).

TABLE 1: DETAILS OF UCI DATASETS USED

| Dataset | No. of Instances | No. of Features | No. of Classes | Application Domain |
|---|---|---|---|---|
| IonosphereEW | 351 | 34 | 2 | Electromagnetic |
| Lymphography | 148 | 18 | 2 | Biology |
| SonarEW | 208 | 60 | 2 | Biology |
| SpectEW | 267 | 22 | 2 | Biology |
| waveformEW | 5000 | 40 | 3 | Physics |
| WineEW | 178 | 13 | 3 | Chemistry |
| Zoo | 101 | 16 | 6 | Artificial |

**4.1.2 Case study 2: Handwritten Digit Recognition**

The following case study includes the database descriptions and feature extraction methods of handwritten digit recognition. For these datasets we have used K-Nearest Neighbor (K=5) classifier as an evaluator in the proposed wrapper FS framework.

**4.1.2.1 Database Description**

For the assessment purpose, six handwritten digit databases are used. Of these, five are offline digit databases written in Bangla, Arabic, Telugu, Devanagari and Gurumukhi, and one online digit database written in Assamese. Bangla, Arabic, Telugu and Devanagari digit databases are taken from CMATERdb ("CMATERDB," n.d.). The dataset size of Telugu, Arabic and Devanagari is 3000 (= 10 × 300) i.e. 300 samples per digit class, whereas Bangla database comprises 6000 (= 10 × 600) samples i.e. 600 samples per digit class. On the other hand, Gurumukhi database contains a total of 2300 (10 × 230) samples taken from the work reported in (Sharma & Jain, 2010). The online Assamese digit database is taken from the work developed by Baruah et al. (Baruah & Hazarika, 2015). As online digits are written on digital pads, these are represented as the sequence of two-dimensional co-ordinate points along with the pen up and pen down information. In this work, all the (X, Y) coordinate points are connected consecutively using DDA line drawing algorithm to form a stroke. Considering pen up and pen down information, all the strokes are generated accordingly to obtain of the final digit image. After that each image is normalized to capture the text part only. Sample images per class is shown in Table 2. The dataset contains only 45-digit sample per class which are further augmented by rotating each image by an angle X. Where X takes only integer value and ranges from −5 to 5 degree. This results 500 samples per class that is further used in the feature extraction.

Initially, all the input images are normalized to 32 × 32 dimension to make the data of consistent resolution that facilitates us to implement any feature extraction technique uniformly.

TABLE 1: IMAGES REPRESENTING EVERY CLASS OF BANGLA, ARABIC, TELUGU, DEVANAGARI, GURUMUKHI AND ASSAMESE SCRIPTS HANDWRITTEN DIGIT SAMPLES

| Class Label | Bangla | Arabic | Telugu | Devanagari | Gurumukhi | Assamese |
|---|---|---|---|---|---|---|
| 0 | 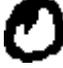 | 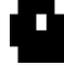 | 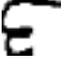 | 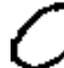 | 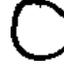 | 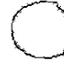 |
| 1 | 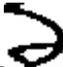 | 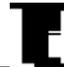 | 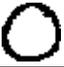 | 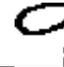 | 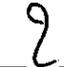 | 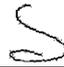 |
| 2 | 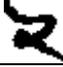 | 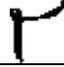 | 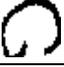 | 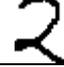 | 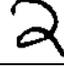 | 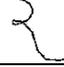 |
| 3 | 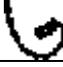 | 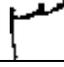 | 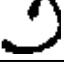 | 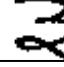 | 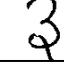 | 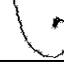 |
| 4 | 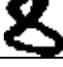 | 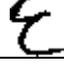 | 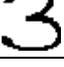 | 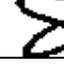 | 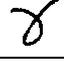 | 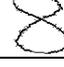 |
| 5 | 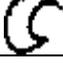 | 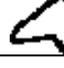 | 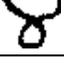 | 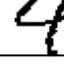 | 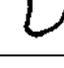 | 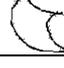 |
| 6 | 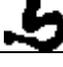 | 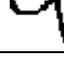 | 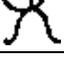 | 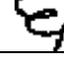 | 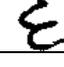 | 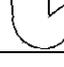 |
| 7 | 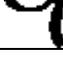 | 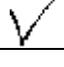 | 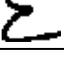 | 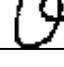 | 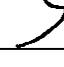 | 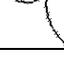 |
| 8 | 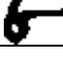 | 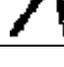 | 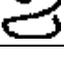 | 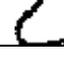 | 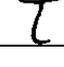 | 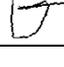 |
| 9 | 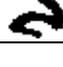 | 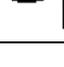 | 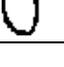 | 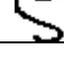 | 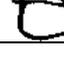 | 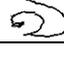 |

**4.1.2.2 Feature Extraction**

In this section, we have adopted Histogram of Oriented Gradients (HOG) for extracting features from the handwritten digit samples. HOG is a texture-based feature descriptor which adopts histogram of the gradients as a statistical measure. Primary idea behind the concept is that, any local shape and object can be demonstrated using the gradient intensity distribution or edge direction (S. Ghosh, Bhowmik, Ghosh, Sarkar, & Chakraborty, 2019). As HOG is invariant to geometric transformation, it is widely used in pattern recognition domain. In the computation part, first of all the entire image is divided into cells. After that, gradients are calculated according to equation 22.

In this work, one dimensional gradient is taken. A matrix $M = \begin{bmatrix} 1 & 0 & -1 \end{bmatrix}$ is used to calculate the gradient in the X direction, $Grad_X$. Subsequently, $M^T$ is adopted for gradient in Y direction, $Grad_Y$. The final gradient direction $Grad_{Dir}$ is calculated using the equation 22.

$$Grad_{Dir} = \tan^{-1} \frac{Grad_Y}{Grad_X} \tag{22}$$

In the further processing, the entire gradient direction domain (0 to 360 degree) is divided into 8 bins. For each cell, the histogram of the obtained gradient directions is calculated based on the 8 bins. Finally, those histograms from each cell are concatenated to get the final feature vector.

**4.1.3 Case Study 3: Text / Non-text Classification:**

In the following two subsections, database description and the feature extraction methods of text non-text classification are explained. We have used Random Forest (no. of trees = 300) classifier for these datasets.

**4.1.3.1 Database Description:**

The text non-text classification database is constructed through the competition called Recognition of Documents with Complex Layouts (RDCL) by International Conference on Document Analysis and Recognition (ICDAR) community. In our present work, 70 pages from RDCL 2015 (Antonacopoulos, Clausner, Papadopoulos, & Pletschacher, 2015) and 75 pages from RDCL 2017 (Clausner, Antonacopoulos, & Pletschacher, 2017) are taken into consideration. The datasets are generated in PRImA research lab, University of Salford, UK. One of the reasons for considering this dataset is that it contains diverse challenges and complexities in terms of non-texts like equations, tables, images, graphic separator etc. Text and non-text components are manually cropped from these pages to generate two sets – texts and non-texts. We have considered total 670 images, where 345 are texts and other 345 are non-texts. All the cropped text non-text images are converted into grey scale image from their RGB format for feature extraction. Some of the examples of text and non-text data are shown in Table 3.

TABLE 3: SAMPLE IMAGES FROM TEXT NON-TEXT RDCL DATABASE USED IN THE PRESENT WORK. (A) – (C) REPRESENT TEXT DATA, WHEREAS, (D) – (F) REPRESENT NON-TEXT DATA.

| 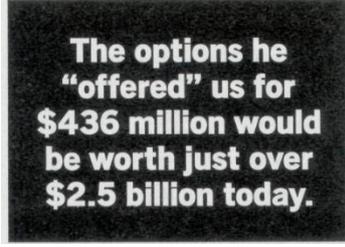 | 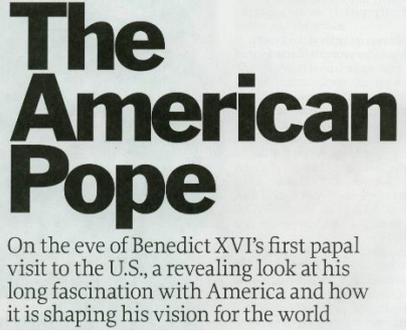 | 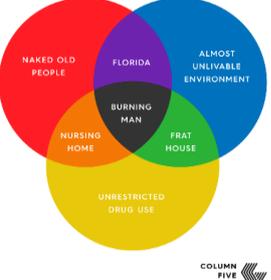 |
|---|---|---|
| (a) | (b) | (c) |
| 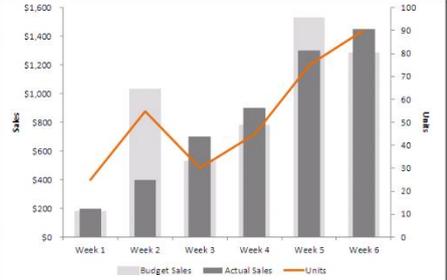 | 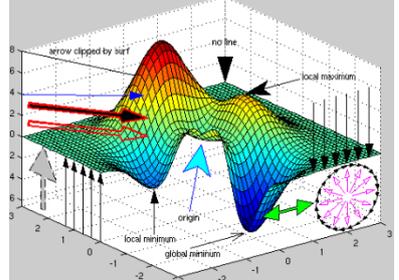 | |
| (d) | (e) | (f) |

### 4.1.3.2 Feature Extraction

In the present work, we have considered five texture-based feature extraction methods – two based on local binary pattern (LBP), one based on local ternary pattern (LTP), and one based on local tetra pattern (LTRP). Firstly, basic concept of LBP is explained followed by LTP, LTRP and their improved versions – uniform and rotation invariant.

**Local Binary Pattern (LBP):** LBP was first proposed by Ojala et al. in their paper (Ojala, Pietikäinen, & Mäenpää, 2002). LBP is a very useful texture-based feature descriptor which tends to find the edge property efficiently. It has been used for text non text classification in the recent past. The detailed formulation of LBP is described here.

Suppose $P_{cen} = X_{cen}, Y_{cen}$ be the center pixel and it is surrounded by $P_{surr}$ number of pixels in a radius R. Each surrounding pixel can be described according to equation 23.

$$(X_i, Y_i) = \left(X_{cen} + R\cos\left(\frac{2\pi i}{P_{surr}}\right), Y_{cen} - R\sin\left(\frac{2\pi i}{P_{surr}}\right)\right) \quad (23)$$

LBP feature ($LBP_{Feature}$) is calculated using those surrounding pixel by comparing their intensity values. If the surrounding pixel is greater than equal to the center pixel, then we assign 1 else 0 is assigned. The formula is given in equation 24.

$$LBP_{Feature} = [LBPf(P_0, P_{cen}), LBPf(P_1, P_{cen}), \ldots\ldots, LBPf(P_{P_{surr}-1}, P_{cen})] \quad (24)$$

The function LBPf is a comparator which is explained in equation 25.

$$LBPf(x, y) = \begin{cases} 1, \text{if } x \geq y \\ 0, \text{if } x < y \end{cases} \quad (25)$$

In our proposed work, we have considered R = 1. As a result, the LBP feature will be a binary string of 8 bit which is further converted to its equivalent decimal value. The final LBP value is formulated in equation 26.

$$LBP_{val}(P_{cen}, P_1, \ldots\ldots, P_{P_{surr}-1}) = \sum_{n=0}^{P_{surr}-1} LBPf(P_n, P_{cen}) \times 2^n \quad (26)$$

**Rotation Invariant LBP:** Ojala et al. (Ojala et al., 2002) have also proposed a modified version of LBP known as rotation invariant LBP (RILBP), which is invariant to image rotation. Firstly, the binary pattern in calculated using LBP. Then, the pattern is rotated bit wise with the carry and new patterns are obtained after each rotation. Finally, the binary pattern that has minimum decimal value, is taken as the feature. The process can be devised using the equation 29.

$$\begin{aligned} RILBP_{(P_{surr}-1)}&(P_{cen}, P_1, \ldots\ldots, P_{P_{surr}-1}) \\ &= \min\{Rotate(LBP_{Feature}, i \mid 0 \leq i \leq P_{surr} - 1)\} \end{aligned} \quad (27)$$

Then we calculate the equivalent decimal value of the rotation invariant binary string using equation 26. The centre pixel is replaced with the obtained value. After replacing all the pixels, we get the RILBP image. Finally, we take the histogram of the values to get the final feature vector.

**Local Ternary Pattern (LTP):** In case of LBP, we only consider whether the surrounding pixel is greater or small compared to the centre pixel. But in case of LTP, it considers the greater and smaller both the cases as well as a threshold is set for the comparison (Xiaoyang Tan & Triggs, 2010). LTP is a three valued code containing $0, +1$ and $-1$. LTP is formulated as follows.

$$LTP_{Feature} = [LTPf(P_0, P_{cen}, thr), LTPf(P_1, P_{cen}, thr), \ldots LTPf(P_{P_{surr}-1}, P_{cen}, thr)] \quad (28)$$

Where, LTPf is the function explained in equation 29, $P_i$ is neighboring pixels, $P_{cen}$ is the center pixel and $P_{surr}$ is the number of surrounding pixels.

$$\text{LTPf(K, L, thr)} = \begin{cases} +1, \text{if } K \geq L + \text{thr} \\ 0, \text{if } |K - L| < thr \\ -1, \text{if } K \leq L - \text{thr} \end{cases} \quad (29)$$

After obtaining the three valued code, it is converted to two binary strings containing only 1s and 0s. One binary string is obtained by replacing all the non 1 (i.e. -1 and 0) with 0 and keeping all the 1 as it is. We get hold of the second binary string by replacing all the non -1s (i.e. 1 and 0) with 0 and converting all the -1 to 1. An example of this conversion is shown.

Let the three valued code be $1100(-1)(-1)00$. The LTP string can be decomposed into two binary strings – $11000000$ and $00001100$.

**Rotation Invariant LTP (RILTP):** RILTP is same as RILBP as explained in equation 31. After obtaining the two binary strings, rotation invariant property is applied on both strings to obtain RILTP. Similarly, the center pixel is replaced with the decimal equivalent of the string. Consequently, we shall get two channels of LTP image due to two strings for each center pixel. At the end, histogram of both the channels are taken as the final feature vector.

**Uniform LTP:** A transition in a binary string is defined as the change of 0 to 1 or change of 1 to 0. The strings that contain less than or equal to two transitions are known as uniform strings and others are known as non-uniform strings. Main aim of using uniform pattern is to remove the redundant features and capture the information properly. Uniform property is incorporated with the two strings produced from LTP. In this case, we only consider the uniform strings and others are ignored. Finally, histogram of the uniform strings is considered as feature vector.

**Local Tetra Pattern (LTRP):** LTRP was first introduced by Murala et al. (Murala, Maheshwari, & Balasubramanian, 2012) in the domain of content based image retrieval. As LTRP is capable of capturing edge information prominently, it can be used in text, non-text separation purpose. The methodology of LTRP is described as follow.

Let us consider S be an image segment of dimension $5 \times 5$. To start with, the first order derivative along vertical (i.e. 90°) and horizontal (i.e. 0°) directions are calculated and denoted by $S_\alpha^1(G_c)|\alpha = 0°, 90°$ where

$$S_{0°}^1 = G_{0°} - G_c \quad (30)$$

$$S_{90°}^1 = G_{90°} - G_c \quad (31)$$

Where, $G_{0°}$ denotes the gray level intensity value of the pixel along 0° and $G_{90°}$ indicates the intensity value along 90°. $G_c$ stands for the center pixel. Horizontal and vertical derivatives can be positive or negative. So, we shall get four different combinations due to their sign. The direction for the center pixel is given according to equation 32.

$$\text{Dir}^1(G_c) = \begin{cases} 1, \text{if } S^1_{0°} \geq 0, S^1_{90°} \geq 0 \\ 2, \text{if } S^1_{0°} < 0, S^1_{90°} \geq 0 \\ 3, \text{if } S^1_{0°} < 0, S^1_{90°} < 0 \\ 4, \text{if } S^1_{0°} \geq 0, S^1_{90°} < 0 \end{cases} \tag{32}$$

Similarly, we can get the direction for each of the neighboring pixels. After considering the direction of all the neighboring pixels, we get the second order $\text{LTrP}^2(G_c)$. This is formulated as equation 33.

$$\text{LTrP}^2(G_c) = \{\text{LTRPf}(\text{Dir}^1(G_c), \text{Dir}^1(G_0)), \ldots, \text{LTRPf}(\text{Dir}^1(G_c), \text{Dir}^1(G_{P_{\text{surr}}-1}))\} \tag{33}$$

Where, LTRPf is defined in equation 34.

$$\text{LTRPf}(\text{Dir}^1(G_c), \text{Dir}^1(G_i)) = \begin{cases} 1, \text{if } \text{Dir}^1(G_c) = \text{Dir}^1(G_i) \\ 0, \text{if } \text{Dir}^1(G_c) \neq \text{Dir}^1(G_i) \end{cases} \tag{34}$$

As there are total 8 neighboring pixels, it will be an 8-bit pattern. Now, the string is four values code. For each direction three binary patterns are generated. As there are total four possible directions for the center pixel, total 12 (= 3 × 4) binary patterns will be generated. An example is shown below where the center pixel direction is 4 ($\text{Dir}^1(G_c) = 4$ according to equation 32).

$$B_{\text{Direction}} = \{\text{LTRPf}_2(\text{LTRP}^2(G_c)[i]| i = 0, 1, \ldots \ldots, P_{\text{surr}} - 1)|_{\text{Direction}=1,2,3} \tag{35}$$

$$\text{LTRPf}_2(x)|_{\text{Direction}=\mu} = \begin{cases} 1, \text{if } x = \mu \\ 0, \text{if } x \neq \mu \end{cases} \tag{36}$$

These strings are generated using the derivative direction. The final string is generated from the gradient magnitude. The gradient magnitude is defined in equation 37.

$$M_{G_i} = \sqrt{S^1_{0°}(G_i)^2 + S^1_{90°}(G_i)^2} \tag{37}$$

The gradient magnitude for all the neighbors and center pixels is calculated. Now, the gradient magnitude is compared among the center and neighbor pixels. The generation of the 13[th] binary string is formulated in equation 38.

$$B_{13} = \{F(M_{G_i} - M_{G_c}) \mid i = 0, 1, \ldots \ldots, 7\} \tag{38}$$

Where,

$$F(x, y) = \begin{cases} 1, \text{if } x \geq y \\ 0, \text{if } x < y \end{cases} \tag{39}$$

We get 13 channels of LTRP image due to the 13 different strings for each center pixel. In case of LTP we got 2 channels, whereas, in LBP we got one channel.

**Uniform LTRP (ULTRP):** The uniform property is described in the previous section. In ULTRP, we consider only the uniform strings among those 13 obtained strings. Finally, histogram of those uniform strings is taken to get the final feature vector.

**Rotation Invariant LTRP (RILTRP):** Previously, rotation invariant property is explained. After getting those 13 strings, rotation invariant property is applied on those strings. Finally, the histogram is considered for the final feature vector.

Number of features obtained by each of the above-mentioned methods is shown in Table 4.

TABLE 2: NUMBER OF FEATURES OBTAINED FROM EACH IMAGE USING EACH OF THE FEATURE DESCRIPTORS RELATED TO RILBP, ULTP, ULTRP, RILTP AND RILTRP.

| Method | Number of features |
| --- | --- |
| RILBP | 36 |
| ULTP | $2 \times 59 = 118$ |
| ULTRP | $13 \times 59 = 767$ |
| RILTP | $2 \times 36 = 72$ |
| RILTRP | $13 \times 36 = 468$ |

**4.1.4 Case study 4: Facial Emotion Recognition**

This section contains the detailed explanation of our fourth case study which is FER. First subsection comprises database description followed by the feature extraction method in the later subsection. For these datasets, we have used Multi-Layer Perceptron classifier as an evaluator in the wrapper FS framework

**4.1.4.1 Database description:**

In the present work, two FER datasets are considered, namely – JAFEE and RaFD. The dataset description along with some pre-processing steps are described in the following sections.

**JAFEE:** JAFEE dataset (Lyons, Akamatsu, Kamachi, & Gyoba, n.d.) consist of facial expression of 10 different Japanese females. There are total 7 facial expression classes (6 basic facial expressions which are angry, disgust, happy, fear, sad and surprise, and 1 neutral facial expression). The dataset contains total 213 images and number of samples belonging to each class is not equal. Hence it is augmented to form a balanced dataset containing total 224 (= 32 × 7) images i.e., 32 images per class. The augmentation is done by incorporating Gaussian white noise of constant variance and mean to 11 image samples. Example of JAFEE database is shown in Figure 5.

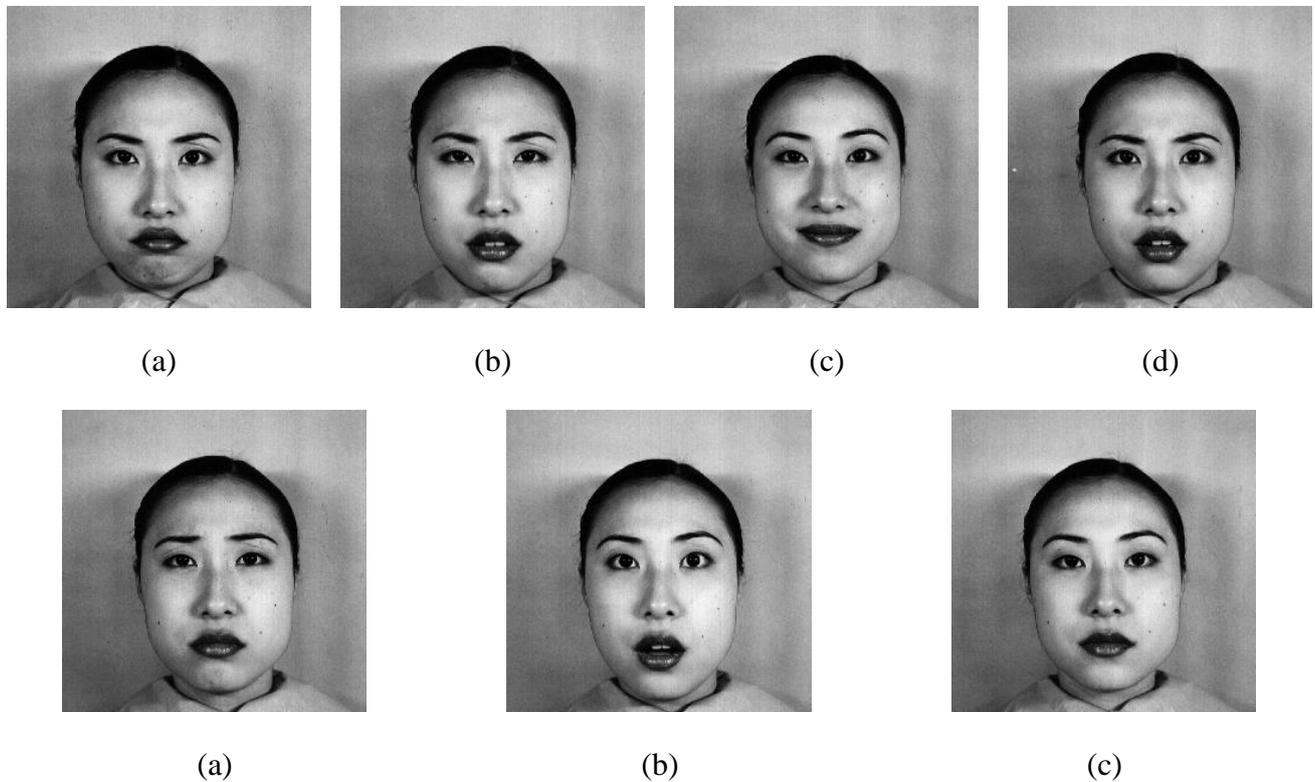

| (a) | (b) | (c) | (d) |

| (a) | (b) | (c) |

Figure 5: Images taken from JAFEE dataset showing various emotions, (a) anger, (b) disgust, (c) happy, (d) fear, (e) sad, (f) surprise and (g) neutral

**RaFD:** RaFD, a facial emotion dataset, (Langner et al., 2010) is collected from 67 models (Caucasian males and females, Moroccan Dutch males, and Caucasian children, both boys and girls), and it contains 8 various expression classes (disgust, fear, happy, neutral, contempt, surprise, sad and angry). Three different gaze directions (frontal, left and right) and five different camera angles are used to capture the facial expressions of each model. Each class contains 201 images, hence the entire dataset size becomes 1608 (= 201 × 8). Figure 6 contains some sample images taken from RaFD database.

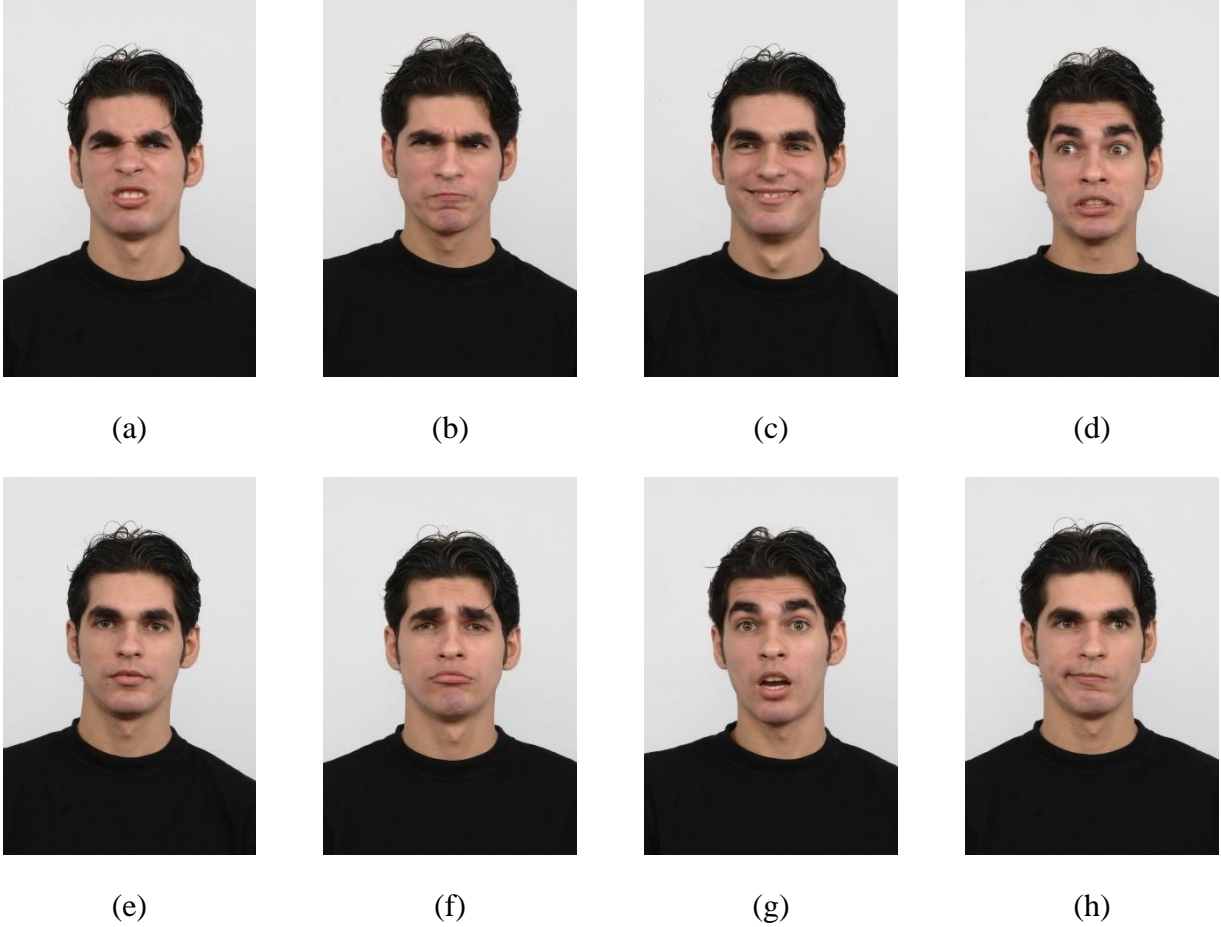

Figure 6: Sample images taken from RaFD dataset displaying various emotions (a) anger, (b) disgust, (c) happy, (d) fear, (e) neutral, (f) sad, (g) surprise and (h) contempt

**Pre-processing:** As facial images are captured in different environmental conditions, hence images may be very noisy, therefore, it hampers the feature extraction process. To overcome this, a proper pre-processing technique is very essential. Viola Jones Haar (Viola & Jones, 2004) has been adopted to capture the important region from the entire image. Point of interest or the important regions mainly contains the area containing facial expressions such as lips, eyebrows, nose, eyes etc. Viola Jones algorithm returns the bounding box coordinates containing the region of interest. To make the method more robust and comparable with real-world scenario, the technique is appraised with various image dimensions. In our work, three image dimensions $32 \times 32, 48 \times 48$ and $64 \times 64$ are considered for the evaluation purpose. After pre-processing, the images are resized to the corresponding resolutions.

### 4.1.4.2 Feature extraction

In the present work, Gabor filter-based feature descriptor which is a frequency-based feature descriptor is used. Gabor filter is a linear filter which is mainly used for texture study. Gabor filter predominantly analyzes if there is any explicit frequency content in the image or not in a precise direction within a localized region around the point or region of interest. Gabor filter is invariant to scale, translation and rotation. Accordingly, it is also robust towards any kind of photometric disorders, mainly occur as illumination changes and image noise (Chengjun Liu & Wechsler, 2002). In the spatial domain, two dimensional Gabor filter involves the sinusoidal plane modulated Gaussian kernel function (Haghighat, Zonouz, & Abdel-Mottaleb, 2015). Equation for the kernel function in the spatial domain is shown in equations (40 - 42). The Gabor features are directly calculated from a gray scale image using these formulas.

$$\text{Gabor}(x,y) = \frac{f^2}{\pi\gamma\eta} \exp\left(-\frac{x'^2 + \gamma^2 y'^2}{2\sigma^2}\right) \times \exp(i.(2\pi f x' + \omega)) \tag{40}$$

$$x' = x \cos\theta + y \sin\theta \tag{41}$$

$$y' = -x \sin\theta + y \cos\theta \tag{42}$$

The standard deviation of the Gaussian envelope is expressed as $\sigma$. $\gamma$ is the spatial aspect ratio and the ellipticity of the support of the Gabor function. i denotes the imaginary number. Phase offset is specified as $\omega$. f stands for Sinusoid frequency and $\theta$ symbolizes the alignment of normal to parallel stripes of the Gabor function. 8 different orientations and 5 distinct scales are taken in the Gabor model that results in 40 diverse Gabor filters.

Multiple spatial resolution and orientation are considered from the set of 2D Gabor filter bank. These are used for convolution of each facial image sample. Let us consider a sample image $\text{img}(x, y)$ and the corresponding Gabor filter kernel is $\Delta_{u,v}(x, y)$. The characterization of sample image $\text{Out}_{u,v}(x, y)$ is formulated according to equation 43 (Ou, Bai, Pei, Ma, & Liu, 2010).

$$\text{Out}_{u,v}(x,y) = \text{img}(x,y) . \Delta_{u,v}(x,y) \tag{43}$$

Finally, the Gabor features are down sampled by a factor of 8. The feature vector size varies with the image dimension. In the present work, image dimensions which are taken into account are - $32 \times 32$, $48 \times 48$ and $64 \times 64$. Table 5 describes the feature dimension corresponds to the previously mentioned image size.

TABLE 5: CALCULATION OF FEATURE VECTOR SIZE FOR VARIOUS IMAGE DIMENSIONS WITH A DOWN SAMPLING FACTOR OF 8. GABOR FILTER BANK CONTAINS 8 DIFFERENT

ORIENTATIONS AND 5 DISTINCT SCALES

| Image Dimensions | Feature Dimension | Feature dimension after down sampling |
|---|---|---|
| $32 \times 32$ | $40960 (= 40 \times 32 \times 32)$ | $640 (= \frac{40960}{8 \times 8})$ |
| $48 \times 48$ | $92160 (= 40 \times 48 \times 48)$ | $1440 (= \frac{92160}{8 \times 8})$ |
| $64 \times 64$ | $163840 (= 40 \times 64 \times 64)$ | $2560 (= \frac{163840}{8 \times 8})$ |

### 4.2 Results

This section describes in detail the results obtained by the proposed methods. First of all, it is required to fix the population size and number of iterations. The number of iterations can be selected through convergence test. The convergence graphs for UCI, Handwritten digit recognition, Text/Non-text and FER datasets are provided in Figures 7-10 respectively. These graphs show how the fitness values of solutions have converged over the iterations for ASOs, ASOs+SA, ASOv and ASOv+SA. By observing these graphs, we have come to a conclusion that all of the four methods converge within 30 iterations. So, for all our experiments, maximum number of iterations has been set as 30.

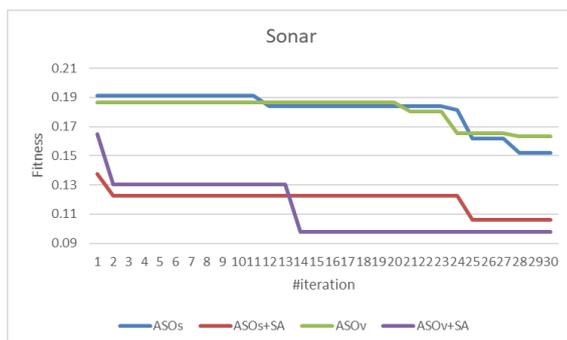
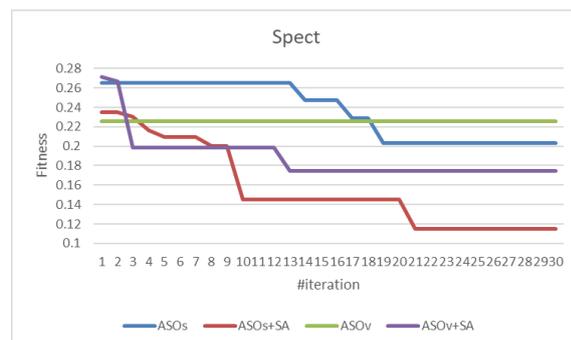

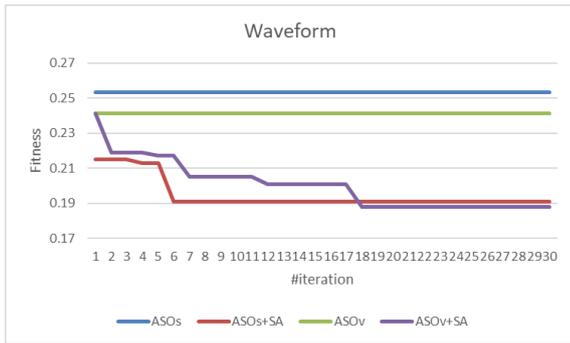
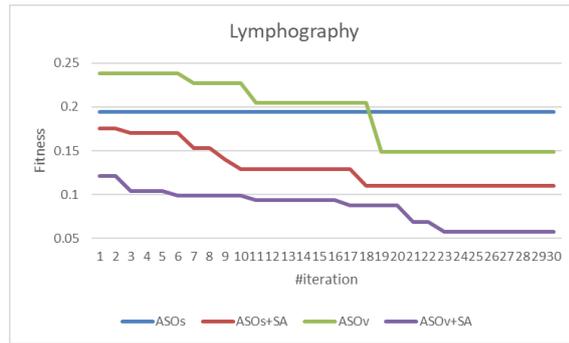
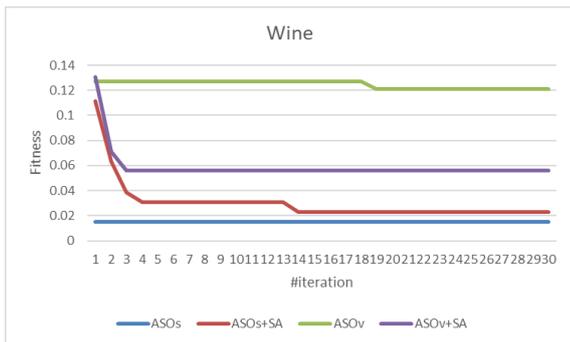
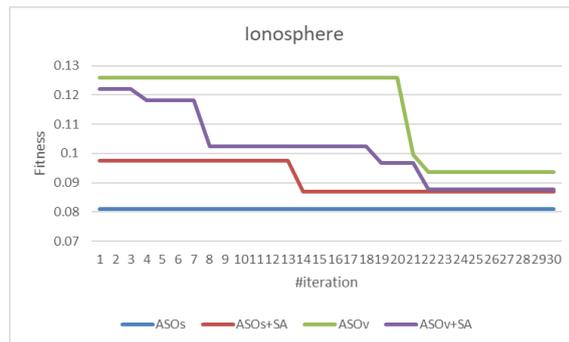
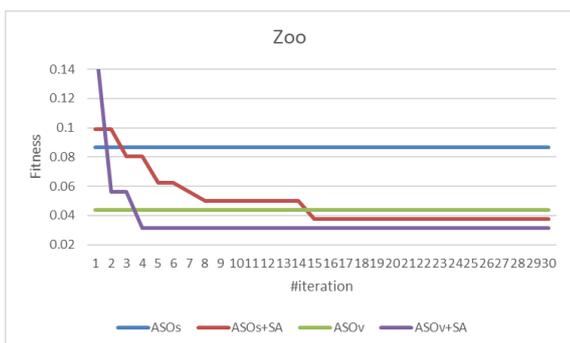
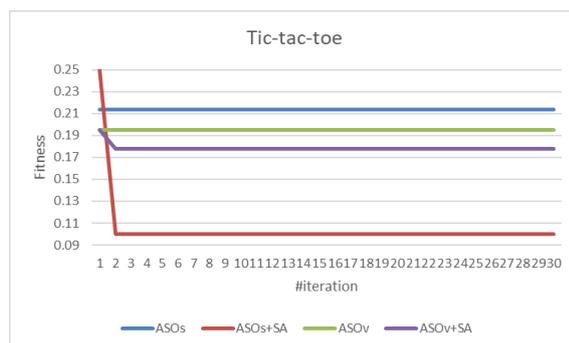

Figure 7: Convergence graph of the proposed FC methods when applied on UCI datasets

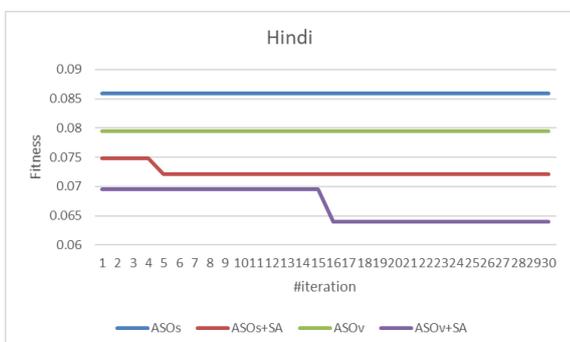
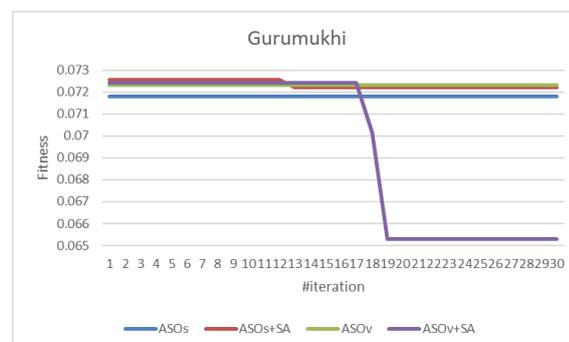

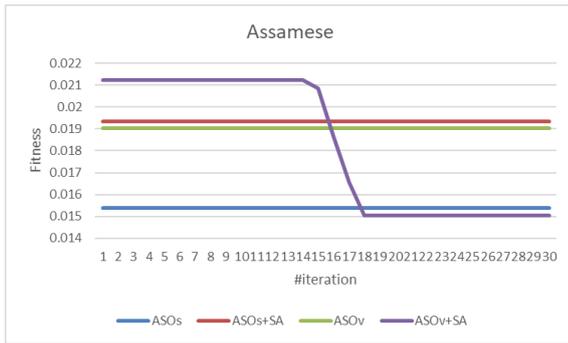
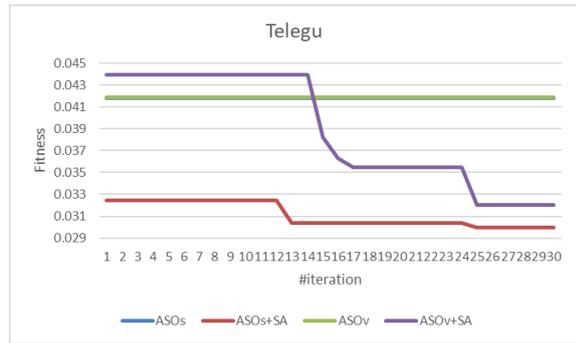
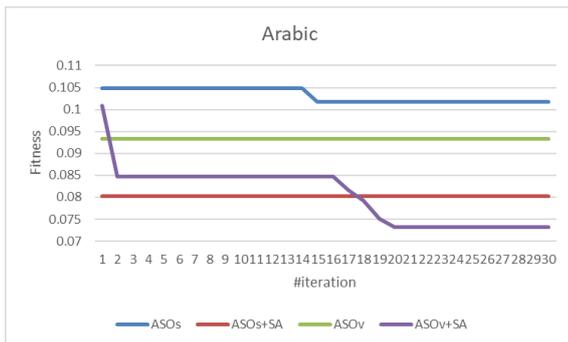
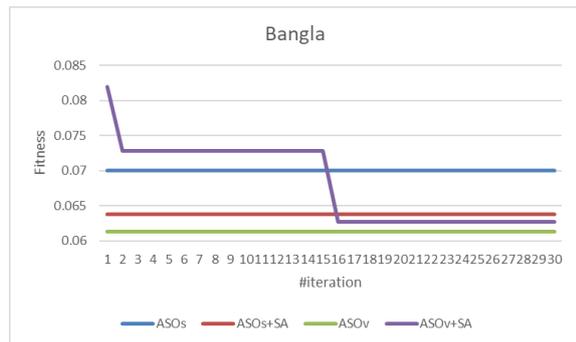

Figure 8: Convergence graph of the proposed FC methods when applied on handwritten digit recognition datasets

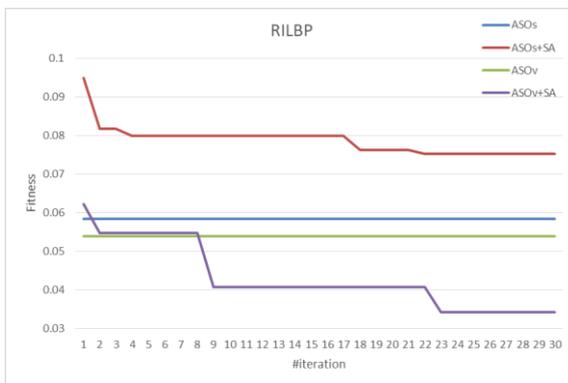
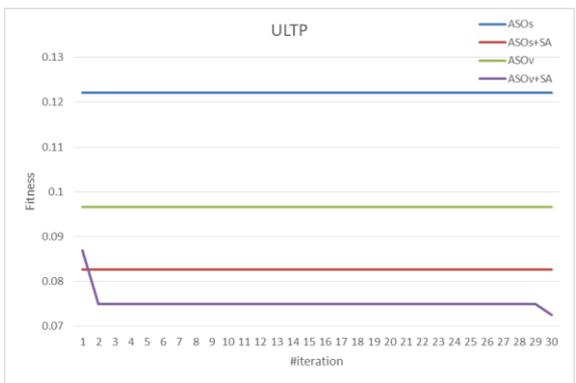
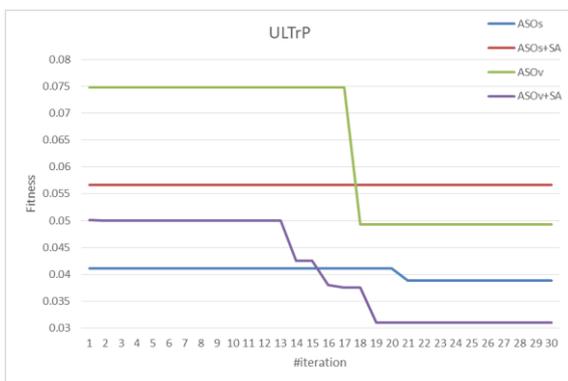
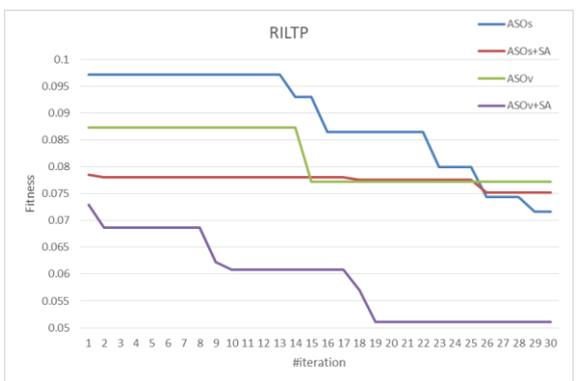

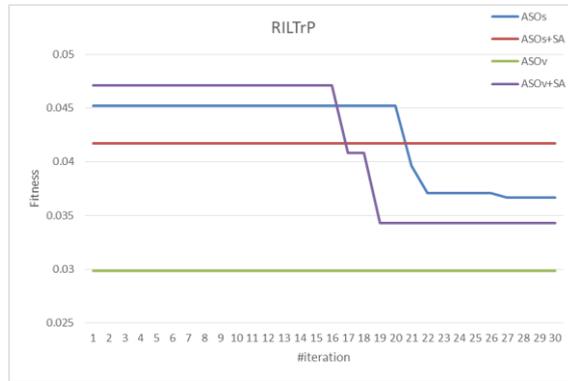

Figure 9: Convergence curve of the proposed approaches for text/non-text separation datasets

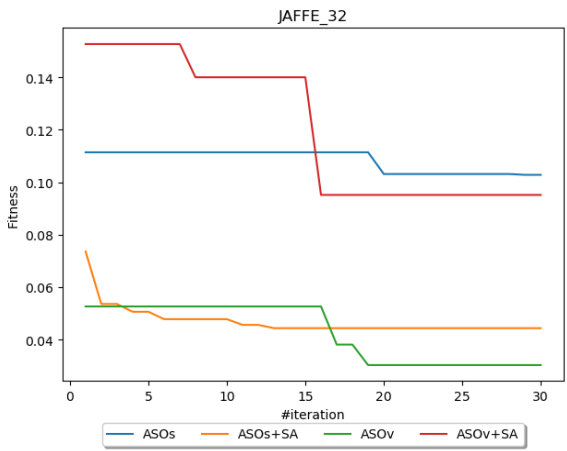
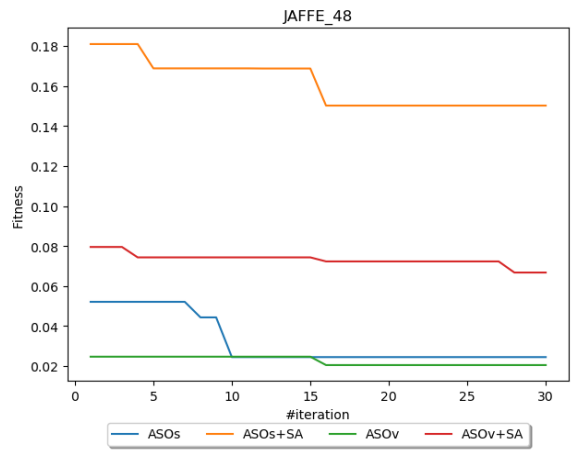
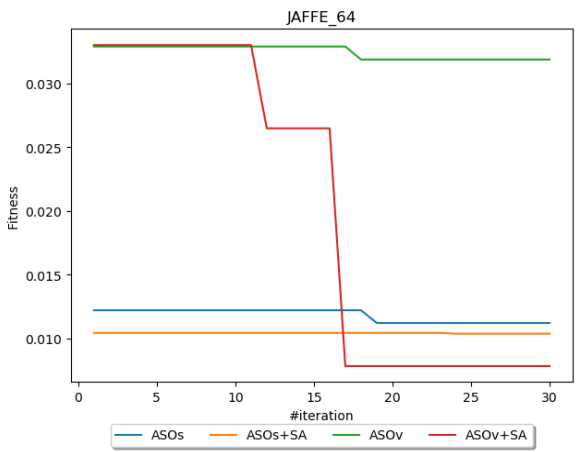
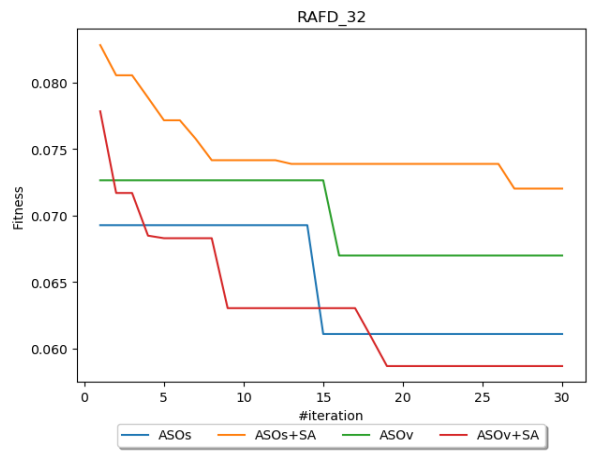

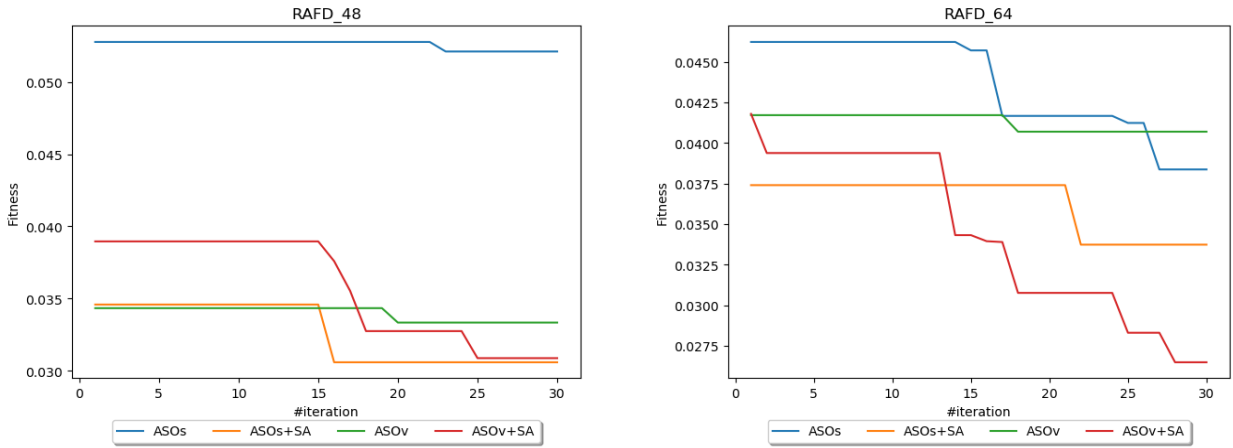

Figure 10: Convergence graph of the proposed approaches for fer datasets

After fixing the number of iterations, we need to get the optimal value for population size. For this, we have again performed several experimentations, and we have plotted classification accuracy vs population size to get the appropriate population size for which the proposed methods achieve the best accuracies over different datasets. The graphs presented in Figures 11-14 depict these classification accuracy vs population size charts over UCI, Handwritten digit, Text/Non-text and FER datasets.

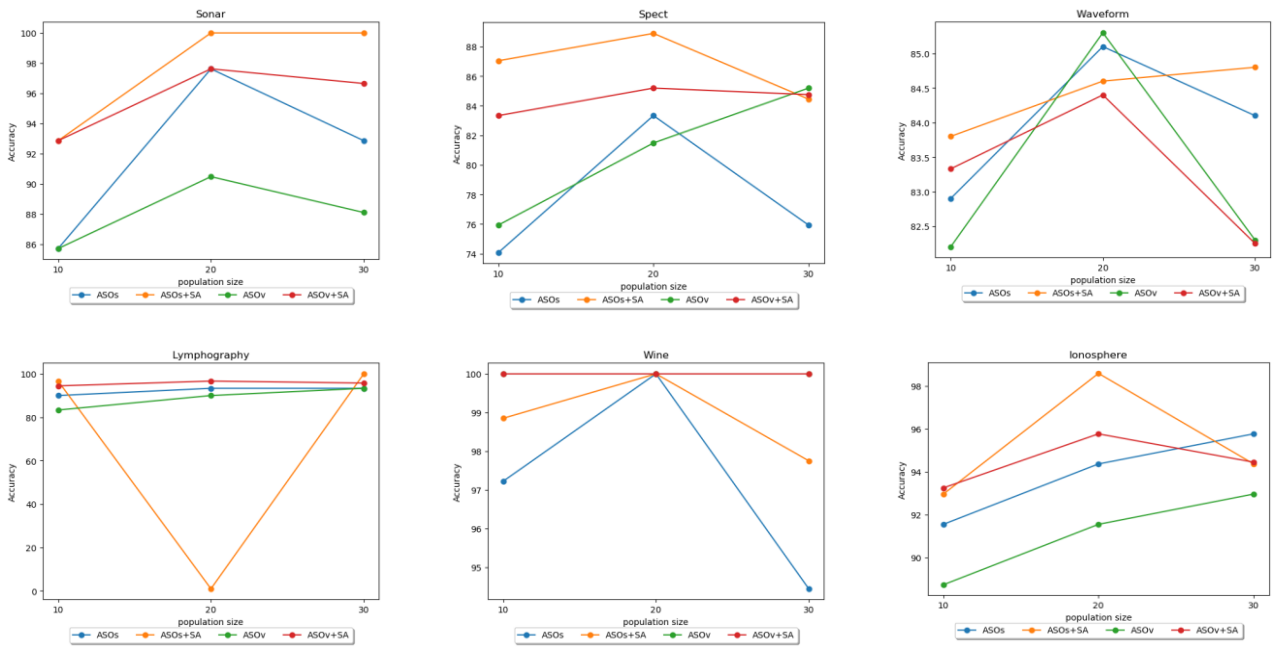

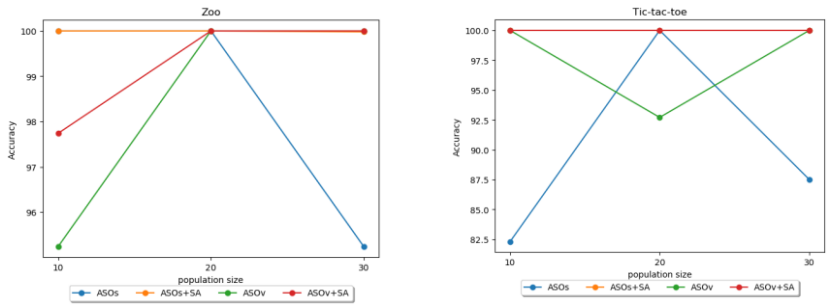

Figure 4: Plot of accuracy vs population size for fixed number of iterations (30) over UCI datasets.

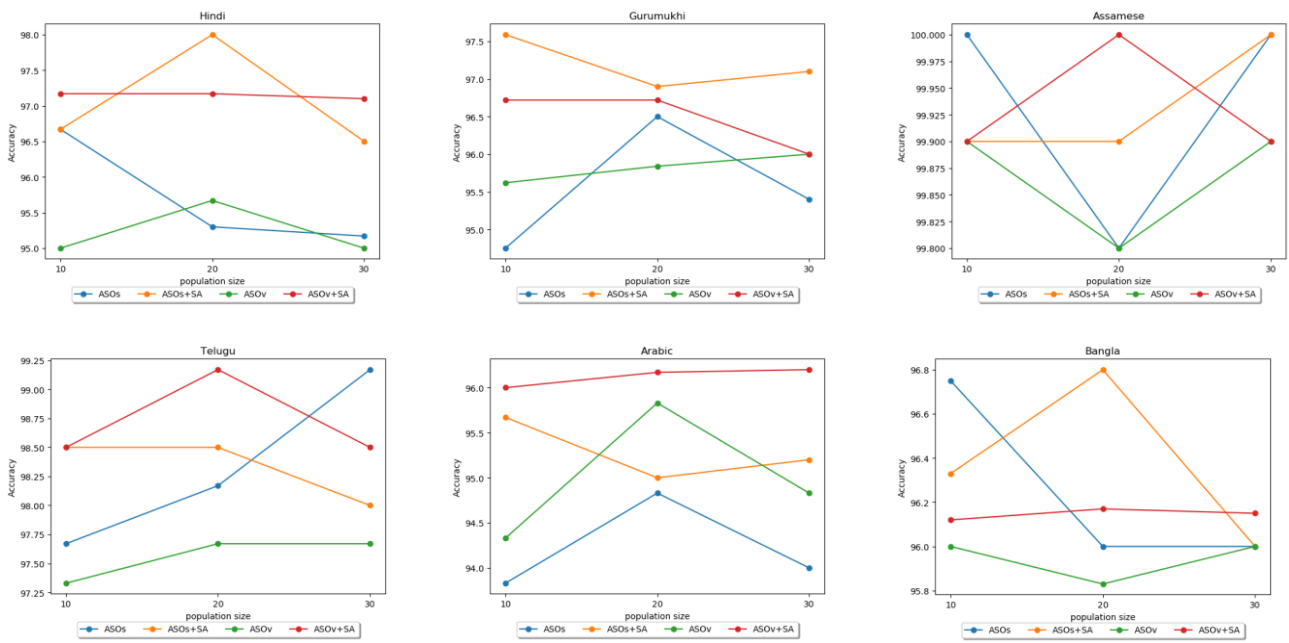

Figure 52: Plot of accuracy vs population size for fixed number of iterations (30) over HANDWRITTEN DIGIT datasets.

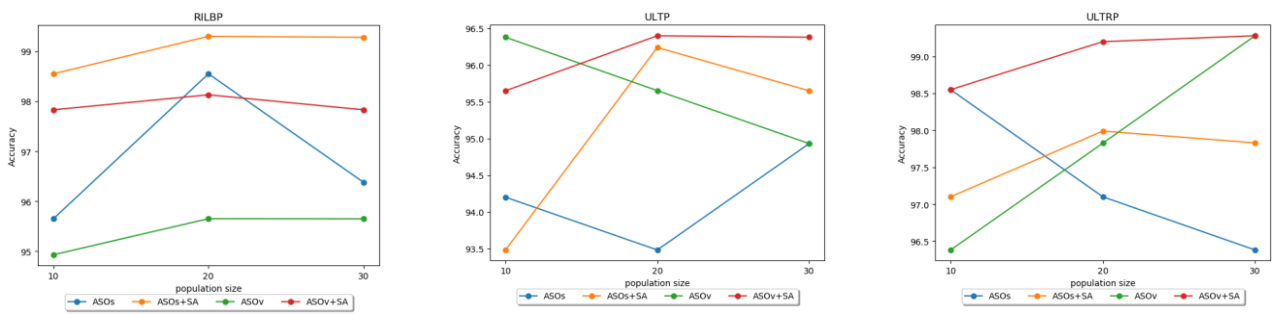

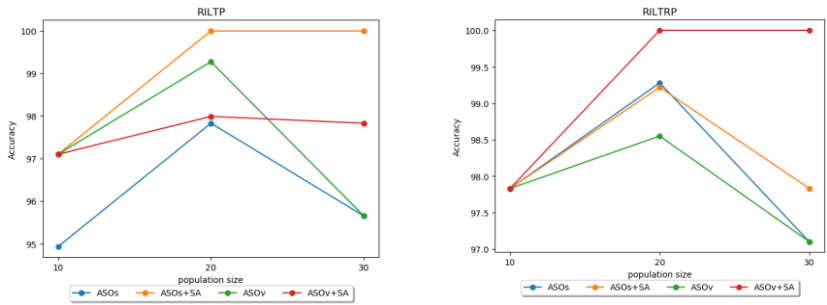

Figure 63: Plot of accuracy vs population size for fixed number of iterations (30) over TEXT/NON-TEXT datasets.

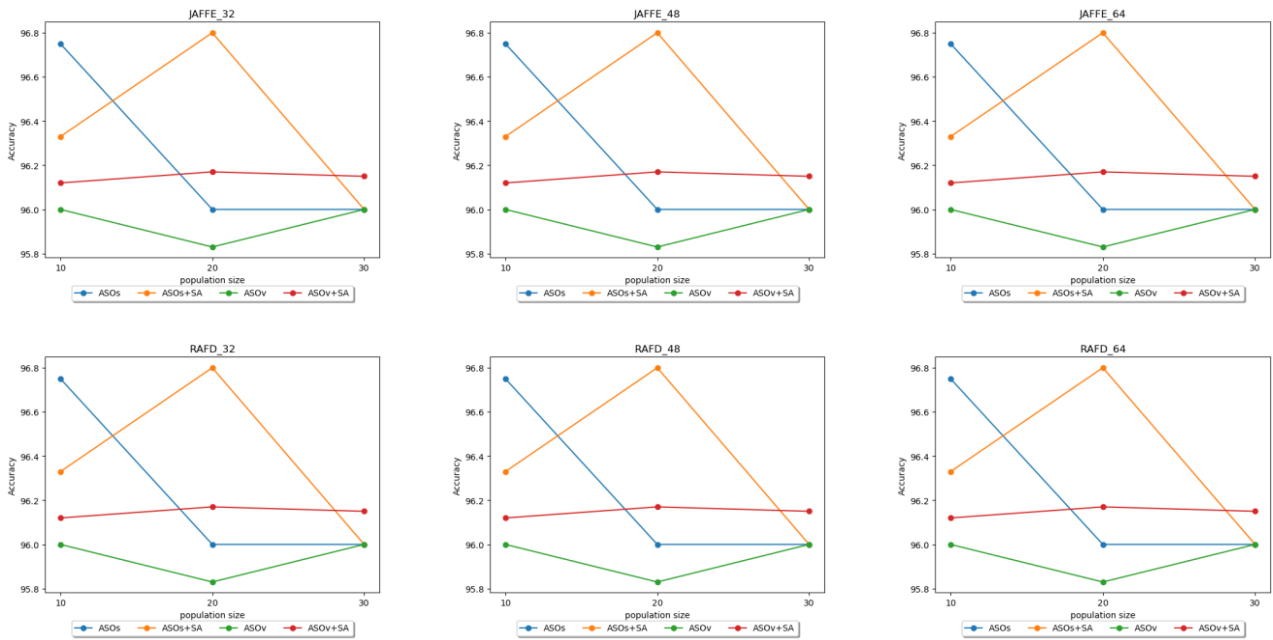

Figure 74: Plot of accuracy vs population size for fixed number of iterations (30) over FER datasets.

### 4.3 Comparison

Since we have applied our methods to various domains, we have compared them with different methods present in the literature align to the corresponding domains. This subsection presents a comparison between our proposed FS methods and other state-of-the-art meta-heuristic FS methods. We have used different classifiers in different cases to show the classifier-independent nature of the proposed methods. For UCI and Handwritten digit datasets, we have used KNN, for Text/Non-text dataset, we have used Random Forest and for FER datasets, we have used MLP classifiers.

#### 4.3.1 UCI Datasets

We have compared the results of our methods obtained over UCI datasets with 6 recent FS methods. For comparison, we have used BGWOPSO (Al-Tashi, Kadir, Rais, Mirjalili, & Alhussian, 2019), BPSO (Al-Tashi et al., 2019), BGA (Al-Tashi et al., 2019), BGOA (M. Mafarja et al., 2019), BGSA (Taradeh et al., 2019), HGSA (Taradeh et al., 2019). The results of BGWOPSO, BPSO and

BGA are obtained from (Al-Tashi et al., 2019) since they have used the same datasets. The results of BGOA, BGSA and HGSA are obtained from (M. Mafarja et al., 2019), (Taradeh et al., 2019) and (Taradeh et al., 2019) respectively. In all the cases, used classifier is KNN (K=5). The accuracies achieved and number of features selected using different methods are shown in Figure 15.

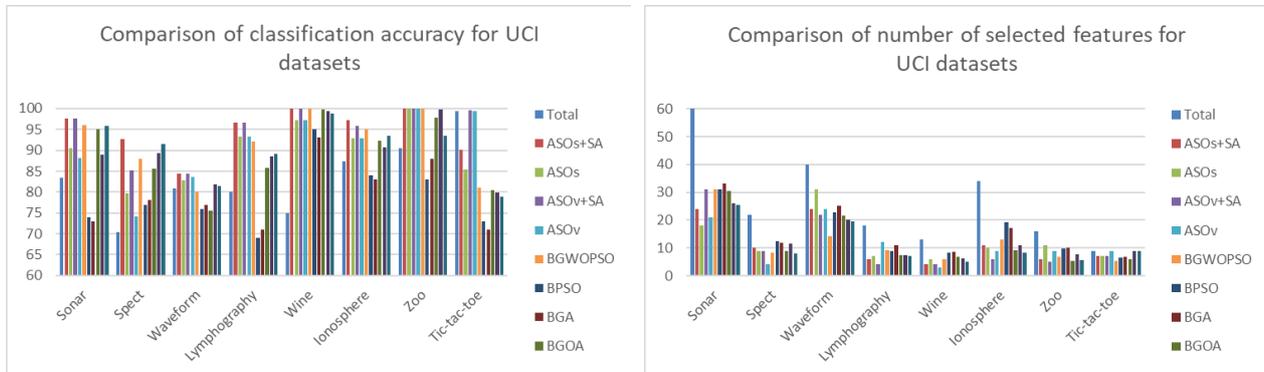

Figure 15: Comparison of our FS methods with some state-of-the-art FS methods in terms of classification accuracy and number of selected features over UCI datasets.

We can easily see that both ASOs-SA and ASOv-SA outperform all the other methods considered here for comparison in terms of classification accuracy. Although, in case of number of selected features, the result is quite different. For most cases, the accuracies achieved by ASOs-SA and ASOv-SA are quite higher than ASOs and ASOv respectively. This justifies the usage of SA as local search. Number of selected features is not always lowest for ASOs-SA or ASOv-SA. In case of waveform and tic-tac-toe datasets, the number of selected features is lowest with BGWOPSO method.

### 4.3.2. Handwritten Digit Recognition Datasets

For handwritten digit recognition datasets, we have compared our methods with BGSA (Esmat Rashedi et al., 2010a), BPSO (Mirjalili & Lewis, 2013), GA (Yang & Honavar, 1998), WFACOFS (M. Ghosh, Guha, Sarkar, & Abraham, 2019). For the other methods, number of agents (number of genes in GA, number of particles in PSO etc.) is taken as 20, since number of atoms is taken as 20 in the proposed methods. Maximum number of iterations has been set to 30 for all the methods. In all cases, used classifier is KNN (K=5). The accuracies achieved and number of features selected using different methods are shown in Figure 16.

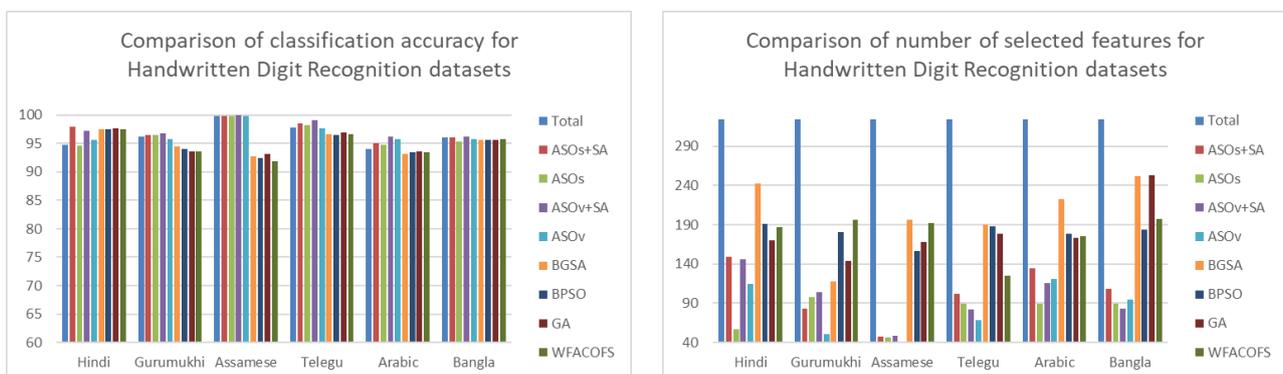

**Figure 8:** Comparison of our FS methods with some state-of-the-art FS methods in terms of classification accuracy and number of selected features for handwritten digit recognition datasets.

Again, ASOs-SA or ASOv-SA outperform all other methods in terms of classification accuracy. Another thing to be noticed is that, ASOs-SA and ASOv-SA perform quite better than ASOs and ASOv respectively. So, we can say that using SA for local search is quite effective. Now, considering the number of selected features, the proposed methods perform significantly better than others. Outperforming all the methods, ASO based methods have selected least number of features. ASOv selects least number of features 3 times, ASOs 2 times and ASOv-SA 1 time.

### 4.3.3 Text/Non-text Classification Dataset

For Text/Non-text classification dataset, we have compared our methods with BGSA (Esmat Rashedi et al., 2010a), BPSO (Mirjalili & Lewis, 2013), GA (Yang & Honavar, 1998), WFACOFS (M. Ghosh, Guha, et al., 2019). For the other methods, number of agents (number of genes in GA, number of particles in PSO etc.) is taken as 20, since number of atoms is taken as 20 in the proposed methods. Number of maximum iterations has been set to 30 for all the methods. In all cases, we have used Random Forest Classifier (number of trees = 300). The accuracies achieved and number of features selected using different methods are shown in Figure 17.

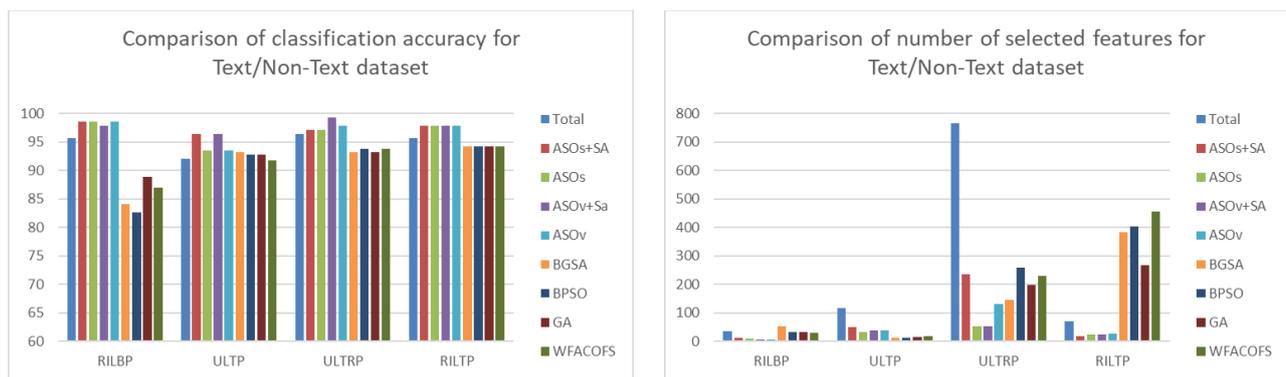

Figure 9: comparison of our FS methods with some state-of-the-art FS methods in terms of classification accuracy and number of selected features for text/non-text separation data.

In this case too, either ASOs-SA or ASOv-SA clearly beats other methods by a significant margin, although the results of ASOs and ASOs-SA are quite closer. The same can be stated for ASOv and ASOv-SA. In case of number of features, except the RILTRP, ASO based methods select lowest number of features. ASOv-SA selects the least number of features 3 times.

### 4.3.4 FER Datasets

For FER datasets, we have compared our method with GA, ME-BPSO (Wei et al., 2017), WOA-CM (M. Mafarja & Mirjalili, 2018) and LHCMA (M. Ghosh, Kundu, Ghosh, & Sarkar, 2019). Amongst those, ME-BPSO, WOA-CM and LHCMA are recently published methods for FS. The accuracies achieved and number of features selected using different methods are shown in Figure 18.

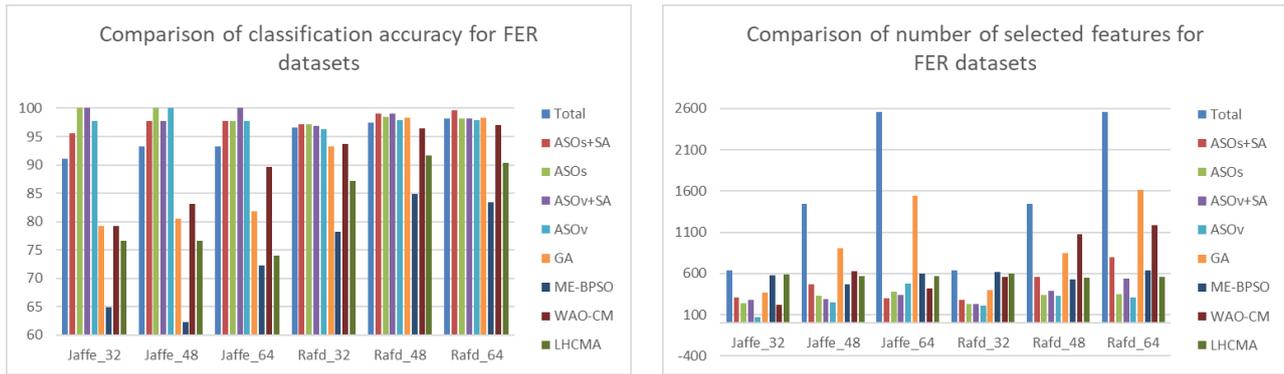

**Figure 10:** Comparison of our FS methods with some state-of-the-art FS methods in terms of classification accuracy and number of selected features for FER datasets.

Considering classification accuracy, in all six cases, ASO based methods have achieved higher accuracy than others. ASOs, ASOs-SA and ASOv-SA have achieved the highest accuracy in multiple cases. In terms of number of features selected, ASOv has performed consistently well. It has selected lowest number of features in 5 cases whereas in another case ASOs has selected lowest number of features.

Considering all the 25 accuracy results, it can be observed that ASO based methods have achieved highest accuracy in all 25 cases. ASOs, ASOs-SA, ASOv and ASOv-SA have achieved highest accuracies respectively in 6, 15, 3 and 16 cases.

Now, considering all the 25 results in terms of number of features selected, ASO based methods have selected lowest number of features in 23 cases i.e., 92% of the cases. ASOs, ASOs-SA, ASOv and ASOv-SA have outperformed others respectively in 5, 2, 9 and 5 cases.

## 5. Conclusion

FS is an important pre-processing technique in enhancing classifiers' ability in classification process. In this paper, we have proposed the FS variant of pre-existing ASO algorithm. While mapping the continuous values in ASO to the binary space of FS, we have used two different transfer functions: one V-shaped (ASOv) and one S-shaped (ASOs). In addition, we have also included SA to enhance exploitation in ASO and named then ASOv-SA and ASOs-SA. To the best of our knowledge, this is the first time ASO has been applied for FS problem.

The performances of the proposed approaches are assessed by applying them on 25 datasets, which are from 4 different categories. We have applied the proposed methods on UCI, Handwritten digit recognition, Text/Non-text separation and FER datasets. These datasets are from completely belong to different domains, which give us clear idea about robustness of the proposed FS techniques. Two criteria are considered, classification accuracy and number of features for measuring the strength of the proposed methods. By using FS, we are able to achieve much higher accuracy with less number of features, irrespective of the domain of the datasets.

To compare the proposed methods with other FS techniques, we have considered different recently published meta-heuristic FS methods. The proposed methods have beat all the methods in comparison in terms of classification accuracy and beat most of the methods in terms of number of selected features. ASOs-SA and ASOv-SA have achieved higher accuracy than ASOs and ASOv respectively, in multiple cases. This justifies the use of SA. Another thing to be noticed, ASOv outperforms ASOs when it comes to number of selected features.

For future studies, we can apply different local search technique with ASO or use different classifiers. Along with that we can try other domains with large feature dimension ($> 10000$). It would also be really interesting if we are able to hybridize ASO with another population based meta-heuristic algorithms.

**Acknowledgement**: This work is partially supported by the CMATER research laboratory of the Computer Science and Engineering Department, Jadavpur University, Kolkata, India, PURSE-II and UPE-II, projects. Ram Sarkar is partially funded by DST grant (EMR/2016/007213). Authors are also thankful to Showmik Bhowmik, Manosij Ghosh, Dipayan Ghosh and Tuhin Kundu for their help to prepare some datasets.

**Conflict of Interest:** Authors declare that there is no conflict of interest.